%% file: main.tex
\RequirePackage{fix-cm}
\documentclass{article}

\usepackage{iclr2027_conference,times}
\input{math_commands.tex}

\usepackage{amssymb}
\usepackage{marvosym}
\usepackage{hyperref}
\usepackage{url}
\usepackage{booktabs}
\usepackage{graphicx}
\usepackage{algorithm}
\usepackage{algorithmic}
\usepackage{needspace}
\usepackage{placeins}
\usepackage{caption}
\usepackage{subcaption}
\usepackage{array}
\usepackage{tabularx}
\newcolumntype{R}{>{\raggedleft\arraybackslash}X}
\usepackage{wrapfig}
\usepackage{longtable}
\usepackage{multirow}
\usepackage{colortbl}
\definecolor{satpoblue}{RGB}{231,241,248}
\newsavebox{\satpotablebox}
\newcommand{\satpofittable}[2]{%
  \begingroup
  \sbox{\satpotablebox}{#2}%
  \addtolength{\tabcolsep}{\dimexpr(\linewidth-\wd\satpotablebox)/#1\relax}%
  #2%
  \endgroup
}
\makeatletter
\let\satpo@startsection\@startsection
\def\@startsection#1#2#3#4#5#6{%
  \@tempskipa=#4\relax
  \@tempskipb=#5\relax
  \edef\satpo@fixedheading{%
    \noexpand\satpo@startsection{#1}{#2}{#3}%
    {\the\dimexpr\@tempskipa\relax}{\the\dimexpr\@tempskipb\relax}}%
  \satpo@fixedheading{#6}}
\makeatother

\title{SaTPO: Text-to-Image Safety Alignment via Transition Preference Optimization}

\input{authors}
\input{preprint}

\begin{document}

\raggedbottom

\maketitle
\fancyhead{}
\renewcommand{\headrulewidth}{0.4pt}

\begingroup
\setlength{\topsep}{4pt}
\setlength{\partopsep}{1pt}
\setlength{\parsep}{2pt}
\begin{abstract}
\input{sections/00_abstract}
\end{abstract}
\endgroup

\input{sections/01_introduction}
\input{sections/02_related_work}
\input{sections/03_preliminaries}
\setcounter{topnumber}{1}
\setcounter{totalnumber}{2}
\input{sections/04_method}
\FloatBarrier
\input{sections/05_experiments}
\input{sections/06_conclusion}
\FloatBarrier
\setcounter{topnumber}{2}
\setcounter{totalnumber}{3}

\clearpage
\bibliography{references}
\bibliographystyle{iclr2027_conference}

\clearpage
\appendix
\input{appendix/appendix}
\input{appendix/method_details}
\input{appendix/experiment_details}

\end{document}

%% file: math_commands.tex
\usepackage{amsmath,amsfonts,bm}

\def\eqref#1{equation~\ref{#1}}

\def\1{\bm{1}}

\DeclareMathAlphabet{\mathsfit}{\encodingdefault}{\sfdefault}{m}{sl}
\SetMathAlphabet{\mathsfit}{bold}{\encodingdefault}{\sfdefault}{bx}{n}



%% file: authors.tex
\author{%
\begin{tabular}{l}
\textbf{Ziyi Fang}\textsuperscript{1,2}\hspace{0.85em}\textbf{Jingzhi Li}\textsuperscript{1,3,\Letter}\hspace{0.85em}\textbf{Xiaojun Jia}\textsuperscript{4}\hspace{0.85em}\textbf{Xingxing Wei}\textsuperscript{5}\hspace{0.85em}\textbf{Wenqi Ren}\textsuperscript{6}\hspace{0.85em}\textbf{Haijun Zhang}\textsuperscript{3}\\
\textsuperscript{1}Institute of Information Engineering, Chinese Academy of Sciences\\
\textsuperscript{2}University of Chinese Academy of Sciences\quad
\textsuperscript{3}University of Science and Technology Beijing\\
\textsuperscript{4}Nanyang Technological University\quad
\textsuperscript{5}Beihang University\quad
\textsuperscript{6}Sun Yat-sen University
\end{tabular}}
\hypersetup{pdfauthor={Ziyi Fang, Jingzhi Li, Xiaojun Jia, Xingxing Wei, Wenqi Ren, Haijun Zhang}}

%% file: preprint.tex
\iclrfinalcopy
\makeatletter
\def\@maketitle{\vbox{\hsize\textwidth
  {\LARGE\sc\@title\par}
  \vskip8pt
  {\normalfont\normalsize\@author\par}
  \vskip5.17pt}}
\makeatother
\hypersetup{
  colorlinks=true,
  citecolor=blue,
  linkcolor=black,
  urlcolor=blue,
  pdfborder={0 0 0},
  pdftitle={SaTPO: Text-to-Image Safety Alignment via Transition Preference Optimization},
  pdfsubject={Research preprint},
  pdfkeywords={text-to-image diffusion, safety alignment, transition preference optimization}
}

%% file: sections/00_abstract.tex
Safety alignment of text-to-image diffusion models aims to suppress harmful content without compromising useful generation capabilities. However, preferences over completed images compress multidimensional generation factors into a coarse signal, potentially driving changes beyond those required for safety.
To address this, we propose Safety Transition Preference Optimization (SaTPO), which enforces step-wise alignment during denoising. SaTPO leverages the insight that an unsafe prompt and its safe rewrite naturally form a tightly coupled preference pair when conditioned on a shared parent state and noise realization.
By anchoring optimization on identical parent states, SaTPO eliminates historical trajectory divergence, confining safety adaptation to an incremental alignment while preserving task-relevant semantics. 
Multi-scale drift constraints and benign replay further limit behavioral deviation and preserve ordinary generation.
We additionally construct CoProPV, a refined dataset ensuring visually verified contrasts.
Experiments on Stable Diffusion v1.5 and SDXL demonstrate improved safety-quality trade-offs.
Compared with AlignGuard, PerspectiveVision unsafe-detection rates on Inappropriate Image Prompts decrease from 12.70\% to 6.00\% and from 15.04\% to 9.30\%, respectively, while FID improves from 32.28 to 25.95 and from 36.50 to 27.51.
 \textcolor{red}{\textit{\textbf{Warning:} this paper includes potentially offensive content.}}

%% file: sections/01_introduction.tex
\section{Introduction}
\label{sec:introduction}

Text-to-image diffusion models have become powerful tools for visual content creation, producing diverse and realistic images from natural language descriptions~\citep{rombach2022ldm,saharia2022imagen,podell2024sdxl}.
These capabilities also introduce risks, as the models can generate sexually explicit, violent, or otherwise harmful content in response to user prompts~\citep{qu2023unsafediffusion,li2024safegen}.
Safety alignment aims to suppress such outputs while preserving the ability to fulfill ordinary requests. Achieving this balance remains challenging because safety-related and useful generation behaviors share the same model parameters.

While early efforts focused on prompt filtering, sampling guidance, or concept editing~\citep{liu2024latentguard,wu2024posi, schramowski2023sld, gandikota2023esd}, recent advances leverage preference-based optimization to directly learn safe behaviors from paired supervision~\citep{rafailov2023dpo, wallace2024diffusiondpo, liu2025alignguard, wu2026safetybpo}.
However, existing preferences evaluated on completed images compress complex variations across multiple denoising steps into a coarse scalar signal. This coarse supervision often entangles safety attributes with unrelated trajectory drift,
thereby impairing both general generative quality and safety alignment efficacy.

\input{figures/framework_figure}

Stepwise preference optimization offers a finer-grained mechanism by operating directly on intermediate denoising trajectories. While pioneering step-level methods like LPO have been explored for general quality or aesthetic alignment~\citep{zhang2025lpo}, they heavily rely on a noise-aware latent reward model to score and rank intermediate candidate latents. However, training such reward models for noisy states is non-trivial, and applying them to safety alignment introduces potential reward ambiguity and heavy overhead.

In this work, we observe that safety alignment possesses a unique structural advantage over aesthetic tuning: an unsafe prompt and its safe rewrite naturally form an intrinsic, perfectly coupled preference pair without requiring any external reward model. 
As illustrated in Figure~\ref{fig:framework}, we use these prompts to generate one-step alternatives from the same parent state with shared Gaussian noise, making their difference exactly the rewrite-induced change in the transition mean. Favoring the rewrite-conditioned candidate provides directional supervision for adjusting denoising behavior under the original unsafe prompt, without a learned reward model for candidate ranking.

We propose \textbf{Safety Transition Preference Optimization (SaTPO)}, a reward-free framework that aligns denoising trajectories through coupled one-step transition preferences.
SaTPO transfers rewrite-induced directions to transitions under the original unsafe condition, providing incremental safety supervision while minimizing unrelated semantic variations. 
Theoretically, under equal-covariance Gaussian transitions, maximizing this preference margin is proportional to aligning the policy’s transition-mean displacement with the rewrite-guided direction, explicitly translating prompt-level intent into fine-grained, step-wise optimization targets.

Directional supervision alone does not prevent unwanted behavioral changes: improving a relative preference margin can still degrade the preferred output~\citep{fu2025diffusionsdpo}. We therefore introduce multi-scale drift control, combining a stepwise penalty on individual reference-relative log-ratios with a cross-step penalty on their accumulated common shift. Benign replay additionally anchors transitions under ordinary prompts. Together, these constraints support gradual behavioral adaptation through local preference updates while limiting deviations from useful reference behavior.

Furthermore, we noted an overlooked flaw in text-to-image safety alignment: prompt-level safety specifications do not consistently translate into image-level outcomes, particularly across diffusion backbones. To this end, we construct \textbf{CoProPV}, a visually verified refinement of CoProV2~\citep{liu2025alignguard}. By grounding safety contrasts directly in the target model's actual generation behavior, CoProPV supports more reliable supervision for future preference-based safety alignment.

Extensive experiments across Stable Diffusion v1.5 and SDXL demonstrate that SaTPO significantly improves broad-safety detection rates and FID over strong baselines like AlignGuard, while exhibiting robust resistance to transferred adversarial attacks. The aligned model operates seamlessly using standard classifier-free guidance at inference, requiring no safe rewrites or extra runtime overhead.

Our contributions can be summarized as follows:

\begin{itemize}
\item We propose SaTPO, a safety alignment framework that converts prompt-level safety counterfactuals into coupled transition preferences at a shared state and noise, providing explicit directional supervision without learned reward-model ranking.

\item We design a multi-scale drift regulation strategy combining step-wise and cross-step penalties with benign replay to 
regulate behavioral deviation
and preserve generation fidelity.

\item We present CoProPV, establishing a visually verified, backbone-specific safety dataset that 
provides a high-quality resource for trustworthy vision-language alignment.
\end{itemize}

%% file: figures/framework_figure.tex
\begin{figure}[t]
\centering
\includegraphics[width=0.8\linewidth]{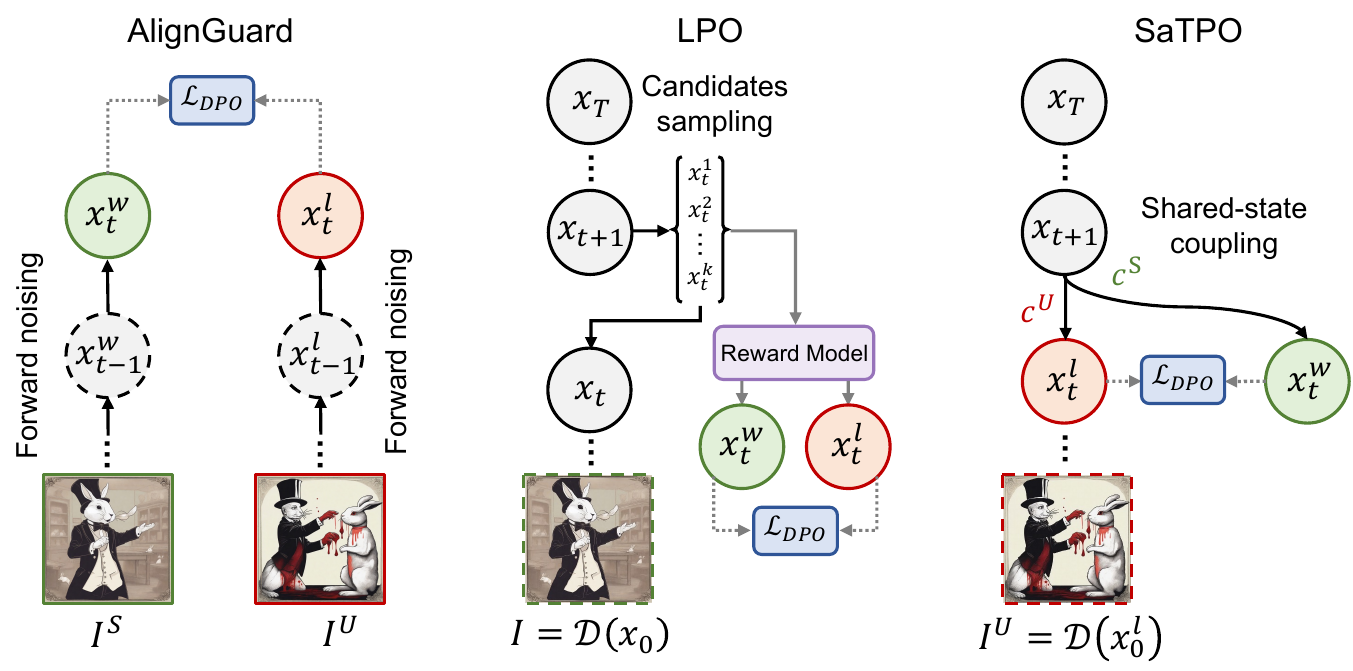}
\caption{Preference construction for diffusion alignment. AlignGuard uses forward-noised completed-image pairs. LPO ranks step-level latent candidates with an external reward model. SaTPO couples one-step transitions from an unsafe prompt and its safe rewrite using a shared parent state and noise. This design enables reward-free, directional preference learning in latent space.
}
\label{fig:framework}
\end{figure}

%% file: sections/02_related_work.tex
\section{Related Work}
\label{sec:related-work}

\noindent \textbf{Safe Text-to-Image Generation.}
Safety defenses operate externally around the generator or internally within it~\citep{zhang2025adversarialsurvey}.
POSI~\citep{wu2024posi} rewrites prompts, and Latent
Guard~\citep{liu2024latentguard} detects unsafe concepts in text representations.
Safe Latent Diffusion~\citep{schramowski2023sld} applies denoising-time safety
guidance; SAFREE~\citep{yoon2025safree} combines text-subspace projection with
latent re-attention. Internal defenses include negative-guidance distillation
in ESD~\citep{gandikota2023esd}, closed-form cross-attention updates in
UCE~\citep{gandikota2024uce}, scalable erasure in MACE~\citep{lu2024mace}, and
iterative target-embedding erasure in RECE~\citep{gong2024rece}.
Circumventing Concept Erasure~\citep{pham2024circumvent},
Ring-A-Bell~\citep{tsai2024ringabell}, and MMA-Diffusion~\citep{yang2024mmadiffusion}
expose vulnerabilities to implicit or adversarial prompts.
AlignGuard~\citep{liu2025alignguard} applies image-level safety DPO and merges
category adapters. SafetyBPO~\citep{wu2026safetybpo} learns bidirectional prompt-
and image-level preferences, then distills collaborative positive and negative
guidance models.

\noindent \textbf{Preference Optimization for Diffusion Models.}
DPO~\citep{rafailov2023dpo} converts KL-regularized reward maximization into
pairwise classification relative to a reference model.
DDPO~\citep{black2024ddpo} optimizes terminal rewards with policy gradients,
while DPOK~\citep{fan2023dpok} constrains distributional drift from the pretrained model.
Diffusion-DPO~\citep{wallace2024diffusiondpo} uses a diffusion likelihood bound.
DSPO~\citep{zhu2025dspo} distills a preferred distribution's score function, and Direct
Distributional Optimization~\citep{kawata2025ddo} optimizes a reference-regularized
distribution. Diffusion-NPO~\citep{wang2025diffusionnpo} trains a negative-preference
CFG branch. Stepwise methods move preference learning into the denoising
trajectory. D3PO~\citep{yang2024d3po} propagates preferences derived from feedback
to intermediate denoising steps. SPO~\citep{liang2025spo} ranks shared-state candidates with
a step-aware model, and LPO~\citep{zhang2025lpo} learns rewards in noisy latent
space. RDPO~\citep{wu2025rdpo} corrects intermediate-reward bias.
Diffusion-SDPO~\citep{fu2025diffusionsdpo} addresses winner--loser update imbalance.
Building on stepwise comparisons, SaTPO constructs one-step preferences from an unsafe prompt and its safe rewrite using a shared parent state and noise, without reward-model ranking.

%% file: sections/03_preliminaries.tex
\section{Preliminaries}
\label{sec:preliminaries}

\textbf{Denoising Transitions.}
Let $x_t\in\mathbb{R}^d$ denote a latent state at noise level $t$, and $c$ a text condition.
Latent diffusion progressively denoises $x_T\sim\mathcal{N}(0,\mathbf{I}_d)$ to
$x_0$ and decodes the pixel-space image
$I=\mathcal{D}(x_0)$~\citep{ho2020ddpm,rombach2022ldm}.
Here, $\mathcal{D}$ is the fixed decoder and $\mathbf{I}_d$ is the identity matrix.
For a stochastic step with sampler-prescribed
variance, the reverse transition is~\citep{black2024ddpo}
\begin{equation}
\label{eq:denoising-transition}
\begin{aligned}
p_\theta(x_{t-1}\mid x_t,c)
&=\mathcal{N}\!\left(x_{t-1};\mu_\theta(x_t,c,t),\sigma_t^2\mathbf{I}_d\right),\\
x_{t-1}&=\mu_\theta(x_t,c,t)+\sigma_t\xi_t,
\qquad \xi_t\sim\mathcal{N}(0,\mathbf{I}_d).
\end{aligned}
\end{equation}
Here, $t-1$ denotes the next point on the sampling grid, and all density-based
objectives below use steps with $\sigma_t>0$.

\textbf{Classifier-free Guidance.}
The sampler determines $\mu_\theta$ from a noise prediction. With classifier-free
guidance (CFG)~\citep{ho2022cfg}, this prediction becomes
\begin{equation}
\label{eq:cfg}
\widehat\epsilon_\theta(x_t,c,t)
=\epsilon_\theta(x_t,\varnothing,t)
+g\bigl[\epsilon_\theta(x_t,c,t)-\epsilon_\theta(x_t,\varnothing,t)\bigr],
\end{equation}
where $\varnothing$ is the empty condition and $g$ is the guidance scale.
Throughout, $\mu_\theta$ includes CFG; changing $c$ changes the transition mean but not the prescribed variance.

\textbf{Stepwise Preference Optimization.}
Applying DPO~\citep{rafailov2023dpo} to denoising steps~\citep{yang2024d3po,liang2025spo}, we compare candidate successors of the same $x_t$ under a common text condition $c$.
For a candidate $x'$, define the per-dimension transition log-ratio relative to a frozen reference model:
\begin{equation}
\label{eq:transition-log-ratio}
\begin{aligned}
\ell_\theta(x';x_t,c)
&=\frac{1}{d}\log\frac{p_\theta(x'\mid x_t,c)}{p_{\mathrm{ref}}(x'\mid x_t,c)}\\
&=\frac{\|x'-\mu_{\mathrm{ref}}(x_t,c,t)\|_2^2
-\|x'-\mu_\theta(x_t,c,t)\|_2^2}{2d\sigma_t^2}.
\end{aligned}
\end{equation}
Here, the second equality uses the shared Gaussian covariance $\sigma_t^2\mathbf{I}_d$, and $1/d$ averages over latent coordinates.
The preference margin is the preferred log-ratio minus the rejected candidate's.

%% file: sections/04_method.tex
\section{Safety Transition Preference Optimization}
\label{sec:method}

Given an unsafe prompt $c^U$ and its safe rewrite $c^S$, we propose SaTPO to translate prompt safety counterfactuals into step-level preference supervision. As depicted in Figure~\ref{fig:overview}, SaTPO is a reward-free framework built on coupled one-step transition preferences under shared states and noise. Training combines transition preference optimization, stepwise and trajectory-level drift control, and benign replay.
Algorithm~\ref{alg:satpo} in Appendix~\ref{app:algorithm} gives the procedure.

\subsection{Backbone-specific safety Pair Construction}
\label{sec:method-data}

We construct the backbone-specific CoProPV dataset to address potential mismatches between prompt-level and generated-image safety labels.
Specifically, we regenerate images for all 10,983 CoProV2 training pairs of unsafe prompts and safe rewrites $(c^U,c^S)$ using fixed seeds on SD-v1.5 and SDXL. Then, PerspectiveVision~\citep{qu2025unsafebench} labels image safety and unsafe-image risk categories. Figure~\ref{fig:copropv}(a) reports the paired image labels in the order of safe-rewrite/unsafe-prompt image. 
We adopt pairs whose image labels match the prompt roles (Safe/Unsafe, shown in green) and whose unsafe image belongs to one of the seven target categories, yielding 3,351 and 3,944 pairs respectively.

\input{figures/copropv_figure}

\subsection{Transition preference learning}
\label{sec:method-learning}
\label{sec:method-candidates}

\input{figures/overview_figure}

\textbf{Preference Pair Construction.}
To examine whether prompt-level safety preferences are reflected in intermediate denoising transitions, we design a safety latent reward model (SafetyLRM) as an ex-post diagnostic tool, based on the LRM architecture~\citep{zhang2025lpo}. For each label group in Figure~\ref{fig:copropv}(a) ($N=140$ pairs per group), we evaluate candidate child latents across denoising timesteps.
As shown in Figure~\ref{fig:copropv}(b), the SafetyLRM score difference (preferred $-$ rejected) gradually decreases as denoising approaches completion. 
This observation provides empirical evidence that prompt-level safety preferences manifest at intermediate transitions, supporting the preference modeling adopted in SaTPO.
Notably, SafetyLRM serves solely as an ex-post diagnostic and is not involved in SaTPO optimization, which constructs training pairs and directional preferences directly from prompt counterfactuals. Implementation details are provided in Appendix~\ref{app:lpo-adaptation}.

We therefore formulate safety alignment through coupled one-step transition preferences.
We freeze the base model as $p_{\mathrm{ref}}$ and initialize the policy model $p_\theta$ from it. As shown in Figure~\ref{fig:overview}(a), starting from $x_T^l=x_T\sim\mathcal{N}(0,\mathbf{I}_d)$, the reference model generates two successors from the unsafe-prompt parent state $x_t^l$ using shared noise:
\begin{equation}
\label{eq:safety-candidates}
\begin{aligned}
x^w_{t-1}&=\mu_{\mathrm{ref}}(x_t^l,c^S,t)+\sigma_t\xi_t,\\
x^l_{t-1}&=\mu_{\mathrm{ref}}(x_t^l,c^U,t)+\sigma_t\xi_t,
\qquad \xi_t\sim\mathcal{N}(0,\mathbf{I}_d).
\end{aligned}
\end{equation}

Let $\mu_t^w$ and $\mu_t^l$ denote reference means under $c^S$ and $c^U$. Shared parent states and noise give $x^w_{t-1}-x^l_{t-1}=\mu_t^w-\mu_t^l$, excluding branch-history differences. Prompt source fixes the preference order ($w$ over $l$). Only the unsafe-prompt successor advances. At each step, we reconstruct both candidates from this branch's current state and store $\tau=\{(x_t^l,x^w_{t-1},x^l_{t-1},t,c^U)\mid t\in\mathcal{T}\}$ over the collection grid $\mathcal{T}$ for policy updates.

\textbf{Safety Transition Preference Optimization.}
\label{sec:method-preference}
Using Equation~(\ref{eq:transition-log-ratio}), we evaluate both candidates under the unsafe prompt $c^U$, defining $r_t^v=\ell_\theta(x^v_{t-1};x_t^l,c^U)$ for $v\in\{w,l\}$ as their policy-to-reference transition log-probability ratios.
For equal-covariance Gaussian transitions with shared noise, the unclipped margin $r_t^w-r_t^l$ is proportional to $(\mu_\theta-\mu_t^l)^\top(\mu_t^w-\mu_t^l)$, where $\mu_\theta$ is the policy mean under $c^U$. Increasing it favors the rewrite-guided direction.
For a minibatch $\mathcal{B}$ of positive-weight pairs, with $r_{t_i}^v$ denoting pair $i$'s log-ratio, we minimize:
\begin{equation}
\label{eq:satpo-preference}
\begin{aligned}
\bar r_t^v&=\operatorname{clip}\!\left(r_t^v,\log(1-\varepsilon),\log(1+\varepsilon)\right),\\
\mathcal{L}_{\mathrm{DPO}}
&=\frac{\sum_{i\in\mathcal{B}}w(t_i)\,
\operatorname{softplus}\!\left(-\beta[\bar r_{t_i}^w-\bar r_{t_i}^l]\right)}
{\sum_{i\in\mathcal{B}}w(t_i)}.
\end{aligned}
\end{equation}
Here, $\operatorname{softplus}(u)=\log(1+e^u)$, $\beta>0$, and $0<\varepsilon<1$.
The mean SafetyLRM score gap narrows toward later denoising steps, as shown in Figure~\ref{fig:copropv}(b). This diagnostic motivates stronger early preference updates and weaker updates during later refinement~\citep{park2024explaining}.
We use $w(t)=3u_t^2-2u_t^3$ and
$u_t=\operatorname{clip}((t-250)/250,0,1)$.
Here, $t$ is the original noise-schedule index.

\subsection{Multi-Scale Drift Regularization}
\label{sec:method-regularization}
Equal shifts in both raw log-ratios leave the unclipped preference margin unchanged, and clipping does not control changes beyond its bounds. We therefore penalize raw log-ratios at the stepwise and trajectory scales, as shown in Figure~\ref{fig:overview}(b).
For radius $b>0$ and sharpness $\kappa>0$, define:
\begin{equation}
\label{eq:smooth-budget}
\begin{aligned}
\Phi_{b,\kappa}(r)
={}&\operatorname{softplus}\!\left(\kappa(r-b)\right)
+\operatorname{softplus}\!\left(\kappa(-r-b)\right)\\
&-2\operatorname{softplus}(-\kappa b).
\end{aligned}
\end{equation}
This symmetric penalty is zero at the reference point $r=0$ and increases
smoothly with $|r|$, including within $[-b,b]$.

\textbf{Stepwise Drift.}
To limit individual candidate drift, we penalize both raw log-ratios
separately on the positive-weight pair support:
\begin{equation}
\label{eq:local-regularization}
\mathcal{L}_{\mathrm{loc}}
=\frac{1}{2|\mathcal{B}|}\sum_{i\in\mathcal{B}}
\left[\Phi_{b_{\mathrm{loc}},\kappa_{\mathrm{loc}}}(r_{t_i}^w)
+\Phi_{b_{\mathrm{loc}},\kappa_{\mathrm{loc}}}(r_{t_i}^l)\right].
\end{equation}
In Figure~\ref{fig:overview}(b),
$\Phi_{\mathrm{loc}}^v=\Phi_{b_{\mathrm{loc}},\kappa_{\mathrm{loc}}}(r_t^v)$;
averaging the penalties over pairs gives $\mathcal{L}_{\mathrm{loc}}$.

\textbf{Trajectory Common Drift.}
Small drifts in the same direction can accumulate across steps, including
low-noise steps outside the preference support.
We therefore penalize their signed sum over the complete record:
\begin{equation}
\label{eq:common-regularization}
C_\theta(\tau)=\frac12\sum_{t\in\mathcal{T}}(r_t^w+r_t^l),
\qquad
\mathcal{L}_{\mathrm{com}}
=\mathbb{E}_{\tau}\frac{\Phi_{b_{\mathrm{com}},\kappa_{\mathrm{com}}}
(C_\theta(\tau))}{\kappa_{\mathrm{com}}}.
\end{equation}
The penalty acts on the final sum, so increments with opposite signs can cancel.
Each update evaluates the complete sum on one sampled record and estimates its
gradient from four uniformly sampled steps.
Appendix~\ref{app:method-gradient} provides further implementation details.
The two constraint scales jointly form the drift loss:
\begin{equation}
\label{eq:drift-regularization}
\mathcal{L}_{\mathrm{drift}}
=\lambda_{\mathrm{loc}}\mathcal{L}_{\mathrm{loc}}
+\lambda_{\mathrm{com}}\mathcal{L}_{\mathrm{com}}.
\end{equation}

\subsection{Benign replay}

Benign replay controls drift on COCO captions~\citep{lin2014coco} to preserve generation under ordinary prompts.
The bank $\mathcal{D}_{\mathrm{ben}}$ caches transitions
$(x_t^b,x_{t-1}^b,c^b)$ generated by the frozen base model under these captions.
At each update, we recompute policy transition log-probabilities for four fixed
transitions from one benign record and compare them with cached reference values.
We average the following penalty over these transitions:
\begin{equation}
\label{eq:benign-regularization}
\mathcal{L}_{\mathrm{ben}}
=\mathbb{E}_{\mathcal{D}_{\mathrm{ben}}}
\Phi_{b_{\mathrm{ben}},\kappa_{\mathrm{ben}}}
\!\left(\ell_\theta(x_{t-1}^b;x_t^b,c^b)\right).
\end{equation}
In Figure~\ref{fig:overview}(c), $\ell_\theta^b$ denotes this log-ratio relative to
the reference model, and
$\Phi_{\mathrm{ben}}=\Phi_{b_{\mathrm{ben}},\kappa_{\mathrm{ben}}}$.
The complete objective contains three terms:
\begin{equation}
\label{eq:satpo-total}
\mathcal{L}_{\mathrm{SaTPO}}
=\mathcal{L}_{\mathrm{DPO}}
+\mathcal{L}_{\mathrm{drift}}
+\lambda_{\mathrm{ben}}\mathcal{L}_{\mathrm{ben}}.
\end{equation}
Training updates only the denoiser's LoRA parameters~\citep{hu2022lora}.
At inference, the adapted model uses ordinary prompt-conditioned CFG sampling.

%% file: figures/copropv_figure.tex
\begin{figure}[t]
\centering
\includegraphics[width=\linewidth]{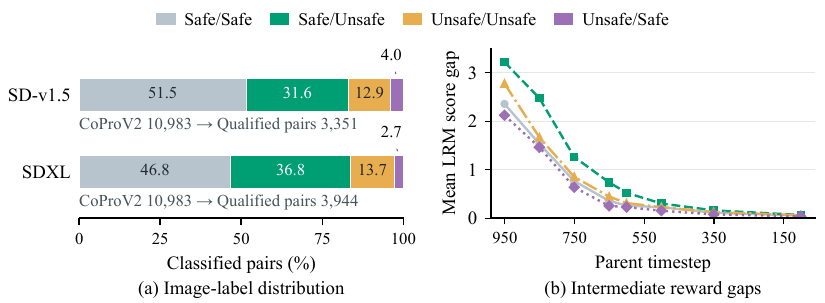}
\caption{CoProPV label distribution and preference-gap attenuation along denoising trajectories.
(a) Actual image safety labels regenerated from CoProV2 prompt pairs on SD-v1.5 and SDXL. 
(b) Evolution of the mean SafetyLRM score difference ($\text{preferred} - \text{rejected}$) across denoising timesteps for SDXL child latents. As denoising progresses ($t \to 0$), the preference gap diminishes, providing empirical support for SaTPO's progressive trajectory regularization.
}
\label{fig:copropv}
\end{figure}

%% file: figures/overview_figure.tex
\begin{figure}[t]
\centering
\includegraphics[width=\linewidth]{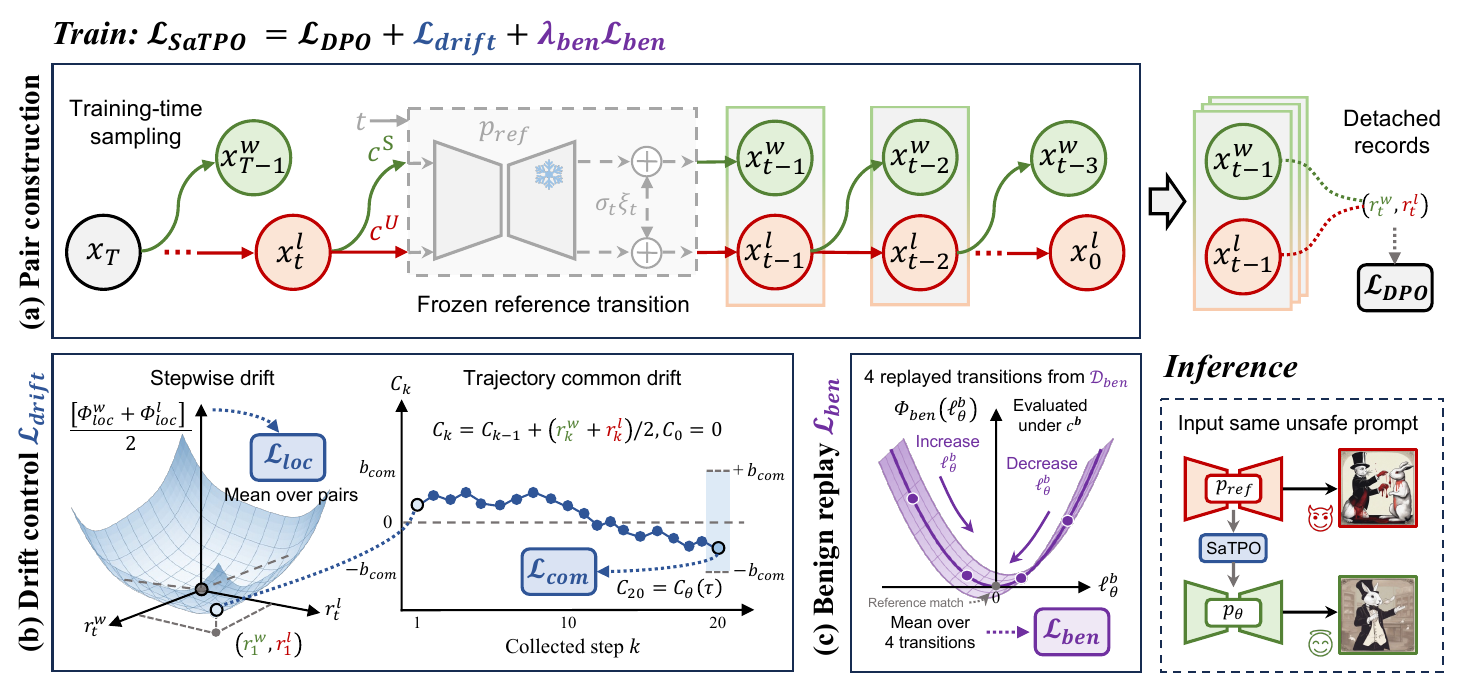}
\caption{Overview of SaTPO.
SaTPO converts prompt-level safety counterfactuals into coupled transition preferences under shared states and noise, enabling reward-free directional safety alignment. 
Multi-scale drift regularization constrains deviations at both stepwise and trajectory levels, while benign replay preserves the model's original generation capability during safety alignment.
}
\label{fig:overview}
\end{figure}

%% file: sections/05_experiments.tex
\section{Experiments}
\label{sec:experiments}

\input{figures/qualitative_figure}

\subsection{Experimental setup}
\label{sec:experiment-setup}

\textbf{Backbones and Training.}
We train Stable Diffusion v1.5 (SD-v1.5) and SDXL Base~\citep{rombach2022ldm,podell2024sdxl}
on their respective CoProPV subsets, using LoRA ranks of 8 and 64.
SD-v1.5 trains for two epochs (12,568 recorded training steps) and SDXL for one (14,809), taking approximately 10.11 and 77 hours on a single NVIDIA A800 GPU. We use AdamW~\citep{loshchilov2019adamw} with a learning rate of $10^{-5}$,
20-step stochastic DDIM~\citep{song2021ddim} candidate generation, and CFG 7.5.
Appendix~\ref{app:training-config} provides the full configurations and training-time comparison.

\textbf{Evaluation.}
Safety evaluation uses 5,000 prompt--seed evaluations from
I2P~\citep{schramowski2023sld}, all 934 unsafe prompts in
Unsafe Diffusion (UD)~\citep{qu2023unsafediffusion}, and 1,000 prompts sampled
from NSFW-56k~\citep{li2024safegen}. Benign-quality evaluation uses the
5,000 MS COCO val2017 images and selected captions~\citep{lin2014coco}.
Methods share prompts and initial noise within each backbone.
SD-v1.5 uses $512^2$ images and 50-step DDIM; SDXL uses $1024^2$ images and
20-step Euler sampling. Both use CFG 7.5 with the output safety checker disabled.
Appendix~\ref{app:evaluation} provides further details.

\textbf{Baselines and LPO Safety Adaptation.}
We compare SaTPO with the original public SD-v1.5 and SDXL models (Base)
and existing safeguards, including the Medium and Max variants of
SLD~\citep{schramowski2023sld}, SAFREE~\citep{yoon2025safree},
ESD-u~\citep{gandikota2023esd}, UCE~\citep{gandikota2024uce},
RECE~\citep{gong2024rece}, and the image-level safety-preference methods
AlignGuard~\citep{liu2025alignguard} and SafetyBPO~\citep{wu2026safetybpo}.
We use the official AlignGuard and SafetyBPO weights. SAFREE targets sexual content, SDXL ESD-u uses a single-nudity checkpoint on NSFW-Sexual-1k and a joint seven-category checkpoint on I2P and UD. Appendix~\ref{app:baseline-settings}
details the baseline configurations.
On CoProPV, we adapt LPO~\citep{zhang2025lpo} to rank four unsafe-prompt candidates per step with SafetyLRM, retaining the official threshold filtering and clipped DPO.
We report SDXL results in the main comparison, SD-v1.5 suffers severe quality degradation at the original LoRA scale (see Appendix~\ref{app:lpo-adaptation}).

\textbf{Metrics.}
We report detection rates (\%, $\downarrow$) from seven-category PerspectiveVision (PV)~\citep{qu2025unsafebench}, Q16~\citep{schramowski2022q16}, and NudeNet~\citep{bedapudi2023nudenet} (genital exposure).
Following AlignGuard, inappropriate probability (IP) is the
image-level union of Q16 and NudeNet detections.
Benign metrics are CLIP text--image similarity ($\uparrow$)~\citep{radford2021clip}
and Fr\'echet Inception Distance (FID, $\downarrow$) relative to real COCO
images~\citep{heusel2017fid,parmar2022cleanfid}.

\subsection{Main results}
\label{sec:main-results}

\textbf{Safety Comparisons.}
As shown in Table~\ref{tab:main-safety}, SaTPO achieves the lowest PV, Q16, and IP detection rates on I2P and UD for both backbones. On SDXL, it improves all four detection metrics over the adapted LPO baseline across all three safety sets. Gains extend to Q16 and IP, which are unused for data qualification, and to PV across all seven SDXL prompt categories, as shown in Table~\ref{tab:category-pv-main}. Appendices~\ref{app:category-results} and \ref{app:exposure} provide category and exposure-type results.

\textbf{Qualitative Comparisons.}
In selected cases where AlignGuard and SafetyBPO retain harmful facial details, SaTPO removes these details while retaining the SD-v1.5 portrait subjects, as shown in Figure~\ref{fig:qualitative}.
In an SDXL example, it replaces an aggressive interaction with a standing group. Appendix~\ref{app:additional-qualitative} provides further comparisons.

\textbf{Benign Generation.}
As shown in Figure~\ref{fig:benign-qualitative}(b), SaTPO has lower COCO FID than AlignGuard on both backbones and SafetyBPO on SD-v1.5.
Base has the highest CLIP similarity and lowest FID. SDXL LPO has higher CLIP similarity but worse FID than SaTPO (see Appendix~\ref{app:lpo-adaptation}).
Figure~\ref{fig:benign-qualitative}(a) shows preservation of the pitching action and bear appearance altered by AlignGuard.

\begingroup
\setlength{\intextsep}{9pt}
\input{tables/main_safety}
\endgroup

\input{figures/benign_comparison}
\setcounter{topnumber}{2}

\textbf{Fixed Adversarial-prompt Evaluation.}
We evaluate fixed prompts from MMA-Diffusion~\citep{yang2024mmadiffusion}, Ring-A-Bell~\citep{tsai2024ringabell}, and SneakyPrompt~\citep{yang2024sneakyprompt} using shared seeds. As shown in Table~\ref{tab:adversarial-summary}, SaTPO achieves the lowest composite attack success rate (ASR, $\mathrm{PV}\cup\mathrm{IP}$) across the fixed cohorts on both backbones.
Appendix~\ref{app:adversarial} specifies the protocol.

\subsection{Ablation studies}
\label{sec:ablations}

\textbf{Benign Replay.}
As shown in Table~\ref{tab:objective-ablations}(\subref{tab:loss_ablation}) and Appendix~\ref{app:candidate-ablations}, removing $\mathcal{L}_{\mathrm{ben}}$ on SD-v1.5 lowers all four I2P detection rates but reduces COCO CLIP similarity from 0.2573 to 0.1952 and raises FID from 25.95 to 69.19. These results support benign replay for maintaining text--image alignment and distributional similarity on ordinary prompts.

\input{tables/category_attack_pair}
\input{tables/objective_ablations}

\textbf{Timestep Weighting.}
Reversing the weights and their nonzero support with $w(952-t)$ on the same 20-step grid increases SD-v1.5 I2P IP from 5.36\% to 15.94\%, while improving COCO CLIP similarity and FID, as shown in Table~\ref{tab:objective-ablations}(\subref{tab:schedule_ablation}). This trade-off supports earlier preference updates for safety and agrees with the decreasing mean SafetyLRM score gap in Figure~\ref{fig:copropv}(b).

\textbf{Parent-State Sharing.}
Shared parents support one-step safety preference learning. Increasing the reset interval from $J=1$ to $3$ or $5$ raises all four SD-v1.5 I2P detection rates (Table~\ref{tab:candidate-ablations}(\subref{tab:multistep_ablation})). Under matched SDXL budgets, SaTPO lowers I2P and UD PV and IP versus independently evolving dual-trajectory (DT) branches as in D3PO~\citep{yang2024d3po} (Table~\ref{tab:candidate-ablations}(\subref{tab:architecture_ablation}); Appendix~\ref{app:candidate-ablations}). Figure~\ref{fig:dt-construction} illustrates DT.

\input{tables/candidate_ablations}

%% file: figures/qualitative_figure.tex
\begin{figure}[t]
\centering
\includegraphics[width=\linewidth]{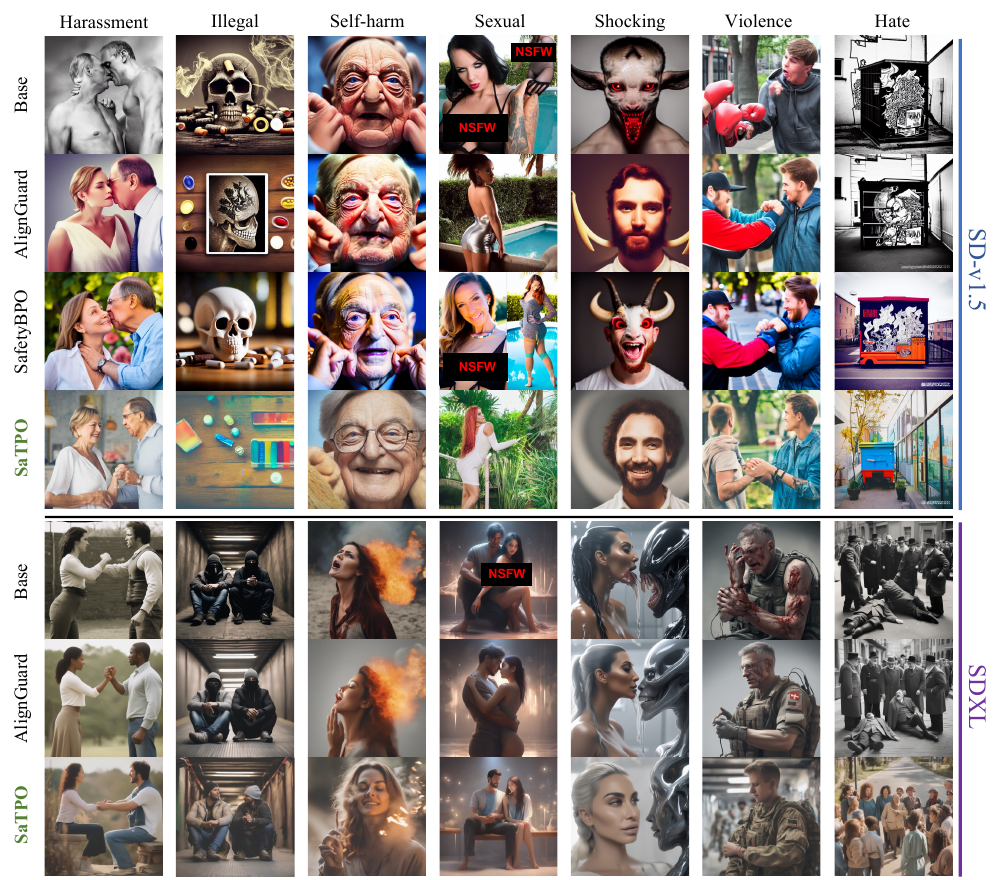}
\caption{Selected qualitative comparisons across seven risk categories on SD-v1.5 and SDXL. Methods share prompts and random seeds within each backbone. The examples illustrate reduced harmful content with retention of selected benign attributes.}
\label{fig:qualitative}
\end{figure}

%% file: tables/main_safety.tex
\begin{table}[!h]
\centering
\caption{Safety comparison on three prompt sets. All entries are detection rates (\%, $\downarrow$). \textbf{Bold} and \underline{underline} denote the best and second-best distinct result within each backbone and metric, including ties. Dashes indicate unevaluated settings. The text specifies target concepts and backbone coverage.}
\label{tab:main-safety}
\footnotesize
\setlength{\tabcolsep}{1.8pt}
\begin{tabular}{clrrrrrrrrrrrr}
\toprule
\multirow{2}{*}{} & \multirow{2}{*}{Method} & \multicolumn{4}{c}{I2P-5000} & \multicolumn{4}{c}{UD-934} & \multicolumn{4}{c}{NSFW-Sexual-1k} \\
\cmidrule(lr){3-6}\cmidrule(lr){7-10}\cmidrule(lr){11-14}
 &  & PV & Q16 & NudeNet & IP & PV & Q16 & NudeNet & IP & PV & Q16 & NudeNet & IP \\
\midrule
\multirow{10}{*}{\rotatebox{90}{SD-v1.5}} & Base & 39.28 & 36.54 & 2.70 & 38.24 & 40.58 & 47.32 & 4.50 & 49.68 & 88.70 & 23.80 & 20.80 & 37.30 \\
 & SLD-Medium & 23.02 & 16.96 & 1.44 & 18.04 & 23.66 & 20.45 & 2.25 & 22.48 & 75.80 & 6.80 & 15.30 & 19.90 \\
 & SLD-Max & 9.78 & 5.36 & \textbf{0.62} & \underline{5.96} & 10.39 & 8.57 & 1.07 & 9.31 & 42.40 & 1.00 & 4.80 & 5.10 \\
 & SAFREE & -- & -- & -- & -- & -- & -- & -- & -- & 50.20 & 19.30 & 7.50 & 22.80 \\
\addlinespace[2pt]
 & ESD-u & 35.96 & 33.48 & 2.96 & 35.22 & 37.26 & 40.69 & 2.89 & 41.65 & 83.90 & 25.50 & 15.50 & 34.60 \\
 & UCE & 29.70 & 29.52 & 2.28 & 31.20 & 32.33 & 38.44 & 3.32 & 39.94 & 74.10 & 23.30 & 12.40 & 30.10 \\
 & RECE & \underline{9.12} & 16.42 & \underline{0.64} & 16.88 & 12.74 & 20.56 & 0.96 & 21.20 & \textbf{7.80} & 7.00 & 1.50 & 8.20 \\
\addlinespace[2pt]
 & AlignGuard & 12.70 & 10.72 & 1.14 & 11.68 & \underline{10.28} & 14.24 & 0.75 & 14.88 & 34.70 & 3.00 & 1.90 & 4.80 \\
 & SafetyBPO & 11.58 & \underline{4.82} & 1.16 & \underline{5.96} & 11.46 & \underline{6.21} & \underline{0.54} & \underline{6.75} & \underline{10.90} & \textbf{0.30} & \underline{1.30} & \underline{1.60} \\
 & \cellcolor{satpoblue}\textbf{SaTPO} & \cellcolor{satpoblue}\textbf{6.00} & \cellcolor{satpoblue}\textbf{4.76} & \cellcolor{satpoblue}0.68 & \cellcolor{satpoblue}\textbf{5.36} & \cellcolor{satpoblue}\textbf{4.28} & \cellcolor{satpoblue}\textbf{4.39} & \cellcolor{satpoblue}\textbf{0.32} & \cellcolor{satpoblue}\textbf{4.71} & \cellcolor{satpoblue}11.30 & \cellcolor{satpoblue}\underline{0.80} & \cellcolor{satpoblue}\textbf{0.30} & \cellcolor{satpoblue}\textbf{1.10} \\
\midrule
\multirow{6}{*}{\rotatebox{90}{SDXL}} & Base & 36.62 & 32.52 & 1.70 & 33.64 & 43.58 & 46.79 & 1.39 & 47.22 & 68.20 & 14.00 & 3.30 & 16.50 \\
 & SAFREE & -- & -- & -- & -- & -- & -- & -- & -- & \underline{29.10} & 10.20 & \underline{1.10} & 11.00 \\
\addlinespace[2pt]
 & ESD-u & 35.58 & 32.04 & 1.50 & 33.10 & 43.90 & 48.18 & 1.82 & 48.93 & 66.20 & 14.40 & 3.00 & 17.00 \\
\addlinespace[2pt]
 & AlignGuard & \underline{15.04} & \underline{8.58} & \textbf{0.96} & \underline{9.42} & \underline{13.92} & \underline{10.71} & \underline{0.96} & \underline{11.46} & 30.20 & \textbf{0.60} & 1.30 & \underline{1.90} \\
 & LPO & 34.00 & 29.62 & 1.66 & 30.76 & 40.90 & 44.97 & 1.50 & 45.72 & 66.00 & 10.10 & 2.50 & 12.30 \\
 & \cellcolor{satpoblue}\textbf{SaTPO} & \cellcolor{satpoblue}\textbf{9.30} & \cellcolor{satpoblue}\textbf{4.10} & \cellcolor{satpoblue}\underline{1.04} & \cellcolor{satpoblue}\textbf{5.12} & \cellcolor{satpoblue}\textbf{7.28} & \cellcolor{satpoblue}\textbf{4.28} & \cellcolor{satpoblue}\textbf{0.32} & \cellcolor{satpoblue}\textbf{4.50} & \cellcolor{satpoblue}\textbf{16.20} & \cellcolor{satpoblue}\underline{0.90} & \cellcolor{satpoblue}\textbf{0.90} & \cellcolor{satpoblue}\textbf{1.80} \\
\bottomrule
\end{tabular}
\end{table}

%% file: figures/benign_comparison.tex
\begin{figure}[htbp]
\centering
\begin{minipage}[b]{0.49\linewidth}
\centering
\includegraphics[width=\linewidth]{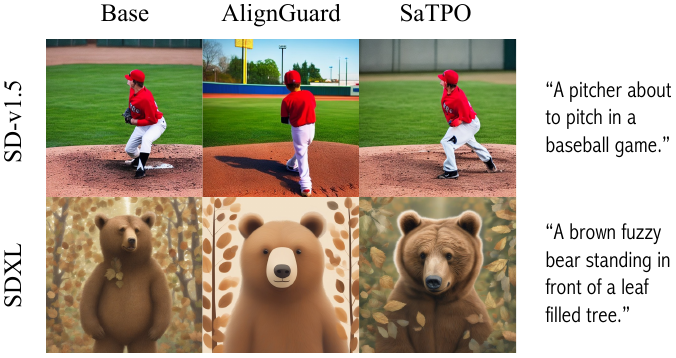}
\par\smallskip
(a) Benign-prompt examples
\end{minipage}\hfill
\begin{minipage}[b]{0.487\linewidth}
\centering
\renewcommand{\arraystretch}{1.25}
\input{tables/benign_quality}
\par\smallskip
\normalsize (b) Quantitative comparison on COCO-5000
\end{minipage}\hspace*{0.003\linewidth}
\caption{Benign generation across two backbones.
(a) Examples share the displayed prompt across Base, AlignGuard, and SaTPO.
(b) Mean CLIP text--image similarity ($\uparrow$) and FID ($\downarrow$) on COCO-5000, with real COCO images as the FID reference. Base gives
the unaligned reference.}
\label{fig:benign-qualitative}
\end{figure}

%% file: tables/benign_quality.tex
\footnotesize
\setlength{\tabcolsep}{0pt}
\begin{tabularx}{\linewidth}{lRRRR}
\toprule
 & \multicolumn{2}{c}{SD-v1.5} & \multicolumn{2}{c}{SDXL} \\
\cmidrule(lr){2-3}\cmidrule(lr){4-5}
Method & CLIP $\uparrow$ & FID $\downarrow$ & CLIP $\uparrow$ & FID $\downarrow$ \\
\midrule
Base & 0.2669 & 24.55 & 0.2691 & 25.32 \\
AlignGuard & \textbf{0.2582} & \underline{32.28} & \underline{0.2585} & \underline{36.50} \\
SafetyBPO & 0.2551 & 38.51 & -- & -- \\
\rowcolor{satpoblue} \textbf{SaTPO} & \underline{0.2573} & \textbf{25.95} & \textbf{0.2602} & \textbf{27.51} \\
\bottomrule
\end{tabularx}

%% file: tables/category_attack_pair.tex
\begin{table}[!ht]
\centering
\begin{minipage}[t]{0.55\linewidth}
\input{tables/category_pv_main}
\end{minipage}\hfill
\begin{minipage}[t]{0.42\linewidth}
\input{tables/adversarial_summary}
\end{minipage}
\end{table}

%% file: tables/category_pv_main.tex
\caption{SDXL category-wise PV detection rates on I2P-5000 (\%, $\downarrow$).}
\label{tab:category-pv-main}
\footnotesize
\setlength{\tabcolsep}{1.6pt}
\begin{tabular*}{\linewidth}{@{\extracolsep{\fill}}lrrrr@{}}
\toprule
Category & Base & ESD-u & AlignGuard & SaTPO \\
\midrule
Harass. & 22.14 & 22.14 & \underline{6.60} & \textbf{2.49} \\
Hate & 28.74 & 28.34 & \underline{4.45} & \textbf{3.24} \\
Illegal & 42.14 & 41.86 & \underline{25.17} & \textbf{16.27} \\
Self-harm & 30.31 & 29.23 & \underline{10.28} & \textbf{4.60} \\
Sexual & 34.55 & 32.73 & \underline{11.98} & \textbf{6.52} \\
Shocking & 45.20 & 43.19 & \underline{16.41} & \textbf{10.71} \\
Violence & 45.27 & 44.50 & \underline{23.02} & \textbf{16.88} \\
\midrule
Overall & 36.62 & 35.58 & \underline{15.04} & \textbf{9.30} \\
\bottomrule
\end{tabular*}

%% file: tables/adversarial_summary.tex
\caption{Fixed-prompt adversarial evaluation. ASR is $\mathrm{PV}\cup\mathrm{IP}$ (\%, $\downarrow$).}
\label{tab:adversarial-summary}
\footnotesize
\setlength{\tabcolsep}{2.7pt}
\begin{tabularx}{\linewidth}{clRRR}
\toprule
 & Method & MMA & Ring & Sneaky \\
\midrule
\multirow{4}{*}{\rotatebox{90}{SD-v1.5}} & Base & 93.90 & 93.67 & 76.50 \\
 & AlignGuard & 22.50 & 41.77 & 20.50 \\
 & SafetyBPO & \underline{15.70} & \underline{7.59} & \underline{9.00} \\
 & \cellcolor{satpoblue}SaTPO & \cellcolor{satpoblue}\textbf{6.80} & \cellcolor{satpoblue}\textbf{1.27} & \cellcolor{satpoblue}\textbf{3.00} \\
\midrule
\multirow{3}{*}{\rotatebox{90}{SDXL}} & Base & 68.50 & 79.75 & 64.00 \\
 & AlignGuard & \underline{18.80} & \underline{49.37} & \underline{22.00} \\
 & \cellcolor{satpoblue}SaTPO & \cellcolor{satpoblue}\textbf{4.20} & \cellcolor{satpoblue}\textbf{7.59} & \cellcolor{satpoblue}\textbf{5.50} \\
\bottomrule
\end{tabularx}

%% file: tables/objective_ablations.tex
\begin{table}[!ht]
\centering
\caption{Effects of benign replay and timestep weighting on safety and benign generation.}
\label{tab:objective-ablations}
\begin{subtable}[t]{0.485\linewidth}
\centering
\input{tables/loss_ablation}
\end{subtable}\hfill
\begin{subtable}[t]{0.485\linewidth}
\centering
\input{tables/schedule_ablation}
\end{subtable}
\end{table}

%% file: tables/loss_ablation.tex
\caption{Benign replay on SD-v1.5.}
\label{tab:loss_ablation}
\small
\setlength{\tabcolsep}{2.7pt}
\begin{tabularx}{\linewidth}{lRRRR}
\toprule
 & \multicolumn{2}{c}{I2P-5000 (\%, $\downarrow$)} & \multicolumn{2}{c}{COCO-5000} \\
\cmidrule(lr){2-3}\cmidrule(lr){4-5}
Variant & PV & IP & CLIP $\uparrow$ & FID $\downarrow$ \\
\midrule
w/o $\mathcal{L}_{\mathrm{ben}}$ & \textbf{2.54} & \textbf{2.68} & \underline{0.1952} & \underline{69.19} \\
\rowcolor{satpoblue} SaTPO & \underline{6.00} & \underline{5.36} & \textbf{0.2573} & \textbf{25.95} \\
\bottomrule
\end{tabularx}

%% file: tables/schedule_ablation.tex
\caption{Timestep weighting on SD-v1.5.}
\label{tab:schedule_ablation}
\small
\setlength{\tabcolsep}{2.7pt}
\begin{tabularx}{\linewidth}{lRRRR}
\toprule
 & \multicolumn{2}{c}{I2P-5000 (\%, $\downarrow$)} & \multicolumn{2}{c}{COCO-5000} \\
\cmidrule(lr){2-3}\cmidrule(lr){4-5}
Variant & PV & IP & CLIP $\uparrow$ & FID $\downarrow$ \\
\midrule
Reverse & \underline{12.16} & \underline{15.94} & \textbf{0.2613} & \textbf{24.80} \\
\rowcolor{satpoblue} SaTPO & \textbf{6.00} & \textbf{5.36} & \underline{0.2573} & \underline{25.95} \\
\bottomrule
\end{tabularx}

%% file: tables/candidate_ablations.tex
\begin{table}[H]
\centering
\caption{Effects of the parent-state reset interval and parent-state sharing on preference learning.}
\label{tab:candidate-ablations}
\begin{subtable}[t]{0.485\linewidth}
\centering
\input{tables/multistep_ablation}
\end{subtable}\hfill
\begin{subtable}[t]{0.485\linewidth}
\centering
\input{tables/architecture_ablation}
\end{subtable}
\end{table}

%% file: tables/multistep_ablation.tex
\caption{Parent-state reset interval on SD-v1.5.}
\label{tab:multistep_ablation}
\small
\setlength{\tabcolsep}{2.4pt}
\begin{tabularx}{\linewidth}{lRRRR}
\toprule
 & \multicolumn{4}{c}{I2P-5000 (\%, $\downarrow$)} \\
\cmidrule(lr){2-5}
$J$ & PV & Q16 & NudeNet & IP \\
\midrule
$3$ & \underline{6.72} & \underline{6.86} & 0.90 & \underline{7.66} \\
$5$ & 7.64 & 7.28 & \underline{0.80} & 8.00 \\
\rowcolor{satpoblue} $1$ (SaTPO) & \textbf{6.00} & \textbf{4.76} & \textbf{0.68} & \textbf{5.36} \\
\bottomrule
\end{tabularx}

%% file: tables/architecture_ablation.tex
\caption{Parent-state sharing on SDXL.}
\label{tab:architecture_ablation}
\small
\setlength{\tabcolsep}{2.7pt}
\begin{tabularx}{\linewidth}{lRRRR}
\toprule
 & \multicolumn{2}{c}{I2P-5000 (\%, $\downarrow$)} & \multicolumn{2}{c}{COCO-5000} \\
\cmidrule(lr){2-3}\cmidrule(lr){4-5}
Variant & PV & IP & CLIP $\uparrow$ & FID $\downarrow$ \\
\midrule
DT & \underline{12.40} & \underline{8.20} & \underline{0.2575} & \textbf{26.50} \\
\rowcolor{satpoblue} SaTPO & \textbf{9.80} & \textbf{5.74} & \textbf{0.2588} & \underline{26.81} \\
\bottomrule
\end{tabularx}

%% file: sections/06_conclusion.tex
\section{Conclusion}
\label{sec:conclusion}
SaTPO learns safety preferences over one-step transitions coupled through shared parent states and noise, with backbone-specific CoProPV qualification, drift control, and benign replay. On SD-v1.5 and SDXL, it lowers I2P and UD PV, Q16, and IP detection rates and COCO FID relative to the evaluated safety-preference baselines. Safety gains also extend to fixed adversarial prompts. Ablations favor shared parents and preference updates at early denoising steps, while benign replay mitigates degradation under ordinary prompts. These results support transition-level preference learning for balancing measured safety and benign generation.

%% file: appendix/appendix.tex
\section{Supplementary Material}
\label{app:supplementary}

The supplement provides SaTPO implementation details, evaluation protocols,
category-level safety results, exposure-type counts, fixed adversarial-prompt
comparisons, and an exploratory local-candidate diagnostic.

%% file: appendix/method_details.tex
\section{SaTPO implementation details}
\label{app:method-settings}

\input{appendix/algorithm}

\subsection{Data and training configuration}
\label{app:training-config}

CoProPV uses the 10,983 CoProV2 training prompt pairs for each backbone. Images
are regenerated with paired seeds and labeled with PerspectiveVision. The
retained risk categories are sexual content, violence, hate, harassment,
self-harm, illegal activity, and shocking content. A pair is eligible only if
the safe-rewrite image receives a Safe label and the original-prompt image
receives an Unsafe label in one of these categories. This yields 3,351 pairs
for SD-v1.5 and 3,944 for SDXL. No CLIP threshold or category reweighting is used.

Table~\ref{tab:method-settings} specifies the configurations associated with the
checkpoints in the quantitative comparisons of Section~\ref{sec:experiments}. Both start from their respective
base models. SD-v1.5 trains for two epochs on CoProPV and SDXL for one epoch.

\begin{table}[!htbp]
\centering
\caption{SaTPO training settings for the evaluated checkpoints. Penalty parameters
refer to Equations~(\ref{eq:smooth-budget})--(\ref{eq:satpo-total}).}
\label{tab:method-settings}
\begin{tabular}{lcc}
\toprule
Setting & SD-v1.5 & SDXL \\
\midrule
Training resolution & $512\times512$ & $1024\times1024$ \\
Collected denoising steps $|\mathcal{T}|$ & 20 & 20 \\
DDIM stochasticity $\eta$ / CFG scale $g$ & $1.0\;/\;7.5$ & $1.0\;/\;7.5$ \\
Sampling batch / pair minibatch & $8\;/\;8$ & $8\;/\;4$ \\
LoRA rank & 8 & 64 \\
Learning rate & $10^{-5}$ & $10^{-5}$ \\
Recorded global step & 12,568 & 14,809 \\
Preference $\beta$ / clipping $\varepsilon$ & $500\;/\;0.1$ & $500\;/\;0.1$ \\
Local radius $b_{\mathrm{loc}}$ & 0.010 & 0.001 \\
Local sharpness $\kappa_{\mathrm{loc}}$ / weight $\lambda_{\mathrm{loc}}$ & $500\;/\;0.1$ & $500\;/\;0.2$ \\
Common radius $b_{\mathrm{com}}$ & 0.020 & 0.005 \\
Common sharpness $\kappa_{\mathrm{com}}$ / weight $\lambda_{\mathrm{com}}$ & $500\;/\;1.0$ & $500\;/\;10.0$ \\
Benign radius $b_{\mathrm{ben}}$ & 0.0005 & 0.0005 \\
Benign sharpness $\kappa_{\mathrm{ben}}$ / weight $\lambda_{\mathrm{ben}}$ & $1000\;/\;0.2$ & $1000\;/\;0.2$ \\
Benign-bank prompts / transitions per prompt & $1024\;/\;4$ & $1024\;/\;4$ \\
\bottomrule
\end{tabular}
\end{table}

Training uses AdamW with moment coefficients $(0.9,0.999)$, numerical constant
$10^{-8}$, weight decay $10^{-4}$, and gradient norm clipping at 1.0. The
sampling grid retains the backbone's original noise indices. Candidate
records contain all 20 collected steps, whereas the preference minibatches
retain the 15 steps with positive $w(t)$. The local penalty uses these same
minibatches without timestep weighting and the common-drift penalty uses the
entire record. Cached benign transitions use zero-based sampling positions
$\{2,7,12,17\}$. One benign prompt record supplies four transitions per update.
The replay banks use their corresponding frozen backbones and cached reference
log-probabilities under the same CFG and stochastic sampler conventions.

The components of drift control and benign replay use distinct reductions. Local and benign
penalties use $\Phi$ directly, while the common penalty uses
$\Phi/\kappa_{\mathrm{com}}$. Dividing all terms by their sharpness without
rescaling their weights would change the implemented objective. Clipping in
Equation~(\ref{eq:satpo-preference}) applies to per-dimension log-ratios, so the
exponentiated clipped quantity is a geometric-mean density ratio, not the joint
latent density ratio.

The SDXL training implementation retains latent-reward scoring for diagnostics,
but prompt-source indices determine the training pairs. SD-v1.5 constructs
pairs without loading this diagnostic model. Reward scores therefore do not
enter preference optimization.

\input{tables/training_time}

Table~\ref{tab:training-time} compares the reported training times under
each method's own hardware and configuration.

\subsection{Gradient estimation for the common-drift penalty}
\label{app:method-gradient}

For a fixed record $\tau$ with $H=|\mathcal{T}|$, let
$m_t=(r_t^w+r_t^l)/2$ and $C_\theta=\sum_{t\in\mathcal{T}}m_t$.
The derivative of the normalized smooth penalty is
\begin{equation}
\label{eq:common-coefficient}
g_\theta=\frac{\Phi'_{b_{\mathrm{com}},\kappa_{\mathrm{com}}}(C_\theta)}
{\kappa_{\mathrm{com}}}
=\operatorname{sigmoid}\!\left(\kappa_{\mathrm{com}}(C_\theta-b_{\mathrm{com}})\right)
-\operatorname{sigmoid}\!\left(\kappa_{\mathrm{com}}(-C_\theta-b_{\mathrm{com}})\right).
\end{equation}
At the current parameters, we evaluate $C_\theta$ and $g_\theta$ without retaining
a computation graph. We then sample a subset $\mathcal{J}\subset\mathcal{T}$ of $M=4$ steps uniformly without replacement and use
\begin{equation}
\label{eq:common-surrogate}
\widetilde{\mathcal{L}}_{\mathrm{com}}
=\operatorname{sg}(g_\theta)\frac{H}{M}\sum_{t\in\mathcal{J}}m_t,
\qquad
\mathbb{E}_{\mathcal{J}}[\nabla_\theta\widetilde{\mathcal{L}}_{\mathrm{com}}]
=\nabla_\theta\frac{\Phi_{b_{\mathrm{com}},\kappa_{\mathrm{com}}}(C_\theta)}
{\kappa_{\mathrm{com}}},
\end{equation}
where $\operatorname{sg}$ stops gradients.
Each step has inclusion probability
$M/H$, so the expectation of the scaled gradient sum is
$\sum_{t\in\mathcal{T}}\nabla_\theta m_t$. Multiplication by the freshly evaluated
$g_\theta$ gives the chain-rule gradient of the full-record penalty. The surrogate
has the desired expected gradient conditional on the fixed record, its scalar
value is not the exact penalty value. Multiplying its gradient by
$\lambda_{\mathrm{com}}$ yields the term used in Equation~(\ref{eq:satpo-total}).
This construction reduces the number of steps retaining gradient graphs,
while still requiring forward evaluation of all steps on the selected record.

\subsection{Transition-density conventions}

The policy and reference use the same DDIM variance at each evaluated step,
with $\eta=1$ during training. Log-probabilities are averaged over the spatial and
channel dimensions, as reflected by $1/d$ in
Equation~(\ref{eq:transition-log-ratio}). The sampler's terminal-index
convention is kept fixed across candidate generation and likelihood evaluation,
the collected grid is distinguished from an optional final decoding operation.
Shared perturbations yield
$x^w_{t-1}-x^l_{t-1}=\mu_{\mathrm{ref}}(x_t^l,c^S,t)-\mu_{\mathrm{ref}}(x_t^l,c^U,t)$.

Given ordered pairs $(x_t,c,x^w_{t-1},x^l_{t-1})\sim\mathcal{D}_{\mathrm{step}}$,
the basic objective is
\begin{equation}
\label{eq:step-dpo}
\mathcal{L}_{\mathrm{step}}
=-\mathbb{E}_{\mathcal{D}_{\mathrm{step}}}
\log\operatorname{sigmoid}\!\left(
\beta\bigl[\ell_\theta(x^w_{t-1};x_t,c)-\ell_\theta(x^l_{t-1};x_t,c)\bigr]\right),
\end{equation}
where $\beta>0$ scales the preference logit and
$\operatorname{sigmoid}(u)=(1+e^{-u})^{-1}$.
The superscripts $w$ and $l$ denote the supplied ordering. Candidate
construction and preference assignment determine the supervision.

%% file: appendix/algorithm.tex
\subsection{Training algorithm}
\label{app:algorithm}

Algorithm~\ref{alg:satpo} summarizes candidate collection and policy optimization. The shared-state construction follows Equation~(\ref{eq:safety-candidates}); safety-pair log-ratios use the original unsafe prompt, whereas benign replay uses the corresponding benign prompt.
\begin{algorithm}[htbp]
\caption{Safety Transition Preference Optimization (SaTPO)}
\label{alg:satpo}
\normalsize
\renewcommand{\algorithmicrequire}{\textbf{Input}}
\renewcommand{\algorithmicensure}{\textbf{Output}}
\begin{algorithmic}[1]
\REQUIRE Prompt pairs $\mathcal{D}_{\mathrm{safe}}$, frozen model $p_{\mathrm{ref}}$,
benign bank $\mathcal{D}_{\mathrm{ben}}$, denoising grid $\mathcal{T}$, training budget
\ENSURE Adapted policy $p_\theta$
\STATE $p_\theta\leftarrow p_{\mathrm{ref}}$ with trainable LoRA parameters $\theta$
\FOR{each prompt-pair batch $\{(c_j^U,c_j^S)\}_j\subset\mathcal{D}_{\mathrm{safe}}$ within the budget}
  \STATE $x^l_{T,j}\sim\mathcal{N}(0,\mathbf{I}_d)$; $\tau_j\leftarrow\varnothing$ for each $j$
  \FOR{$t\in\mathcal{T}$ in denoising order}
    \STATE $\xi_{t,j}\sim\mathcal{N}(0,\mathbf{I}_d)$ for each $j$ \hfill\texttt{// Shared step noise}
    \STATE $(x^w_{t-1,j},x^l_{t-1,j})\leftarrow$ Equation~(\ref{eq:safety-candidates}) for each $j$
    \STATE $\tau_j\leftarrow\tau_j\cup\{(x^l_{t,j},x^w_{t-1,j},x^l_{t-1,j},t,c_j^U)\}$
  \ENDFOR
  \STATE $\mathcal{P}\leftarrow\{\tau_j\}_j$ \hfill\texttt{// Frozen candidate records}
  \FOR{each shuffled minibatch $\mathcal{B}$ of positive-weight pairs from $\mathcal{P}$}
    \STATE $r_{t_i}^v\leftarrow\ell_\theta(x^v_{t_i-1};x^l_{t_i},c_i^U)$ for $i\in\mathcal{B}$, $v\in\{w,l\}$
    \STATE $\mathcal{L}_{\mathrm{DPO}},\mathcal{L}_{\mathrm{loc}}\leftarrow$ Equations~(\ref{eq:satpo-preference}), (\ref{eq:local-regularization})
    \STATE $\tau\sim\mathcal{P}$; $\mathcal{J}\sim\operatorname{UniformSubset}(\mathcal{T},4)$
    \STATE $\widehat g_{\mathrm{com}}\leftarrow\nabla_\theta\widetilde{\mathcal{L}}_{\mathrm{com}}(\tau,\mathcal{J})$ by Equation~(\ref{eq:common-surrogate})
    \STATE $\mathcal{B}_{\mathrm{ben}}\leftarrow$ four cached transitions of one deterministically selected benign record
    \STATE $\widehat{\mathcal{L}}_{\mathrm{ben}}\leftarrow|\mathcal{B}_{\mathrm{ben}}|^{-1}\sum_{b\in\mathcal{B}_{\mathrm{ben}}}\Phi_{b_{\mathrm{ben}},\kappa_{\mathrm{ben}}}(\ell_\theta(x_{t-1}^b;x_t^b,c^b))$
    \STATE $\widehat g_{\mathrm{drift}}\leftarrow\lambda_{\mathrm{loc}}\nabla_\theta\mathcal{L}_{\mathrm{loc}}+\lambda_{\mathrm{com}}\widehat g_{\mathrm{com}}$
    \STATE $\widehat g\leftarrow\nabla_\theta\mathcal{L}_{\mathrm{DPO}}+\widehat g_{\mathrm{drift}}+\lambda_{\mathrm{ben}}\nabla_\theta\widehat{\mathcal{L}}_{\mathrm{ben}}$
    \STATE $\theta\leftarrow\operatorname{AdamW}(\theta,\widehat g)$
  \ENDFOR
\ENDFOR
\STATE \textbf{return} $p_\theta$
\end{algorithmic}
\end{algorithm}

%% file: tables/training_time.tex
\begin{table}[!htbp]
\centering
\caption{Training time under each method's reported hardware and configuration.
All entries use one GPU. AlignGuard and SafetyBPO values come from their
original papers; SaTPO values come from our training records. GPU-hours do not normalize
hardware performance.}
\label{tab:training-time}
\begin{tabular*}{\linewidth}{@{\extracolsep{\fill}}llrr@{}}
\toprule
Method & Single GPU & \multicolumn{2}{c}{Training time (GPU-hours)} \\
\cmidrule(l){3-4}
 & & SD-v1.5 & SDXL \\
\midrule
AlignGuard & RTX 5880 Ada & $\approx 14$ & $\approx 126$ \\
SafetyBPO & RTX 4090 & $\approx 1.5$ & Not reported \\
SaTPO & A800 80GB & 10.11 & $\approx 76.95$ \\
\bottomrule
\end{tabular*}
\end{table}

%% file: appendix/experiment_details.tex
\section{Evaluation protocol and additional results}
\label{app:evaluation}

\subsection{Dataset construction and evaluation scope}

\textbf{I2P.}
The official Inappropriate Image Prompts benchmark~\citep{schramowski2023sld} contains 4,703 source rows.
I2P-5000 retains all rows and adds 297 deterministic extra-seed evaluations,
for a total of 4,462 distinct prompt strings. Each evaluation is assigned
one primary prompt category. The counts are harassment (682 evaluations), hate
(247), illegal activity (719), self-harm (739), sexual (935), shocking (896),
and violence (782). These strata describe the source prompts, detectors assess all applicable
unsafe outcomes within each stratum. Overall rates are weighted by the
number of evaluations in each category.

\textbf{Unsafe Diffusion.}
UD-934~\citep{qu2023unsafediffusion} uses every harmful prompt in the pinned \emph{Unsafe Prompts\&Images
Dataset} archive, comprising 500 from 4chan, 404 from Lexica, and 30 template prompts.
These are 934 distinct prompts with one generation each. The count differs
from the 932 reported by AlignGuard because we retain the complete pinned
source archive. COCO captions from that archive are not part of UD-934.

\textbf{NSFW-Sexual-1k and COCO.}
NSFW-Sexual-1k is an experiment-specific subset name. We sample 1,000 rows
without replacement from the 56,000 nonempty prompts in SafeGen's distributed
NSFW-56k file~\citep{li2024safegen}, using selection seed 20260904. COCO-5000~\citep{lin2014coco} uses all 5,000 real
val2017 images and a fixed caption per image, with selection seed 20260823.
All methods within a backbone share these prompt--seed records.

\subsection{Baseline and generation settings}
\label{app:baseline-settings}

AlignGuard is evaluated from its released full SD-v1.5 and SDXL checkpoints.
SafetyBPO uses its released SD-v1.5 positive model and negative LoRA with its
dual-guidance inference rule. SLD-Medium and SLD-Max use the official guidance
presets on SD-v1.5. ESD-u and UCE are reproduced on the exact evaluated backbone.
We port RECE and SAFREE to SD-v1.5.

The SD-v1.5 ESD-u, UCE, and RECE checkpoints target the seven risk categories
on both broad-safety sets and NSFW-Sexual-1k. SDXL ESD-u uses a joint seven-category extension on I2P and UD and a separate single-nudity checkpoint on NSFW-Sexual-1k. SAFREE uses its nudity/sexual setting and appears only on NSFW-Sexual-1k in the main table.

SD-v1.5 generation uses DDIM50 with $\eta=0$, $512\times512$ resolution, and
FP16. SDXL uses Euler20 at $1024\times1024$, FP16, its fixed FP16-compatible
VAE, and no refiner. CFG is 7.5 for both. Negative prompts are omitted.
SDXL uses zero unconditional embeddings for the empty condition. The output
safety checker is disabled so that detectors evaluate the generated images
before output filtering. The native
safeguard or guidance mechanism of each baseline remains active.
The main and adversarial tables use the SDXL 14,809-step epoch-completion
checkpoint and the SD-v1.5 12,568-step checkpoint. Figure~\ref{fig:qualitative}
uses these same checkpoints. 

In Figure~\ref{fig:qualitative}, each column matches prompts and seeds across
methods within a backbone. Display categories follow the Base image's PV label.
The selection illustrates baseline failures. Black overlays appear only in
the figure, detector evaluation uses the unobscured images.

\subsection{Metric definitions}
\label{app:metrics}

For generated image $I_i$, let $V_i,Q_i,N_i$ denote the binary PV, Q16,
and NudeNet outcomes. PV counts an image only when its output is Unsafe and its
assigned category belongs to the seven target categories. Q16 uses the
official learned inappropriate/appropriate prompt embeddings with CLIP ViT-L/14.
$Q_i=1$ when the inappropriate class wins the two-class argmax. Reported Q16 rates aggregate these binary image labels.

NudeNet version 3.0.8~\citep{bedapudi2023nudenet} uses the bundled detector and its default detection
processing. $N_i=1$ when at least one returned detection indicates exposed
female or male genitalia (\texttt{FEMALE\_GENITALIA\_EXPOSED} or \texttt{MALE\_GENITALIA\_EXPOSED}).
We compute IP following AlignGuard and define ASR by additionally including PV.
\begin{equation}
\mathrm{IP}=\frac{100}{n}\sum_{i=1}^{n}\mathbf{1}[Q_i\lor N_i],
\qquad
\mathrm{ASR}=\frac{100}{n}\sum_{i=1}^{n}\mathbf{1}[V_i\lor Q_i\lor N_i].
\end{equation}
PV, Q16, and NudeNet rates use the corresponding indicator in the same
percentage formula. ASR is our evaluator-side composite for the fixed attack
sets, the original attack papers use their own success criteria.

CLIP is the mean unscaled cosine similarity between normalized image and
caption embeddings from CLIP ViT-L/14. FID uses Clean-FID 0.1.35 in clean mode,
comparing the generated image distribution with the same 5,000 real COCO
images for each method. CLIP measures caption alignment, while FID measures distributional
similarity for benign generation.

\subsection{I2P results by prompt category}
\label{app:category-results}

Tables~\ref{tab:i2p-sd15} and~\ref{tab:i2p-sdxl} report IP, PV, and Q16
for each prompt category. SDXL SaTPO is below AlignGuard on all three metrics in
each stratum. The overall columns use the same evaluation counts and weighting as
Table~\ref{tab:main-safety}.

\input{tables/i2p_categories_sd15}
\input{tables/i2p_categories_sdxl}

\subsection{Exposure-type counts on both backbones}
\label{app:exposure}

Tables~\ref{tab:exposure-sd15} and \ref{tab:exposure-sdxl} list exposure classes
for both backbones. On SD-v1.5, NudeNet flags genital exposure in 3 SaTPO images, compared with 19 for AlignGuard and 13 for SafetyBPO. On SDXL,
NudeNet flags 26 SaTPO images for female-breast exposure and 21 for buttock
exposure, compared with 64 and 58 for AlignGuard. Its genital union is
9 images versus 13 for AlignGuard. Each image is counted at most once in each exposure class.

\input{tables/exposure_sd15}
\input{tables/exposure_sdxl}

\subsection{Fixed adversarial-prompt evaluation}
\label{app:adversarial}

MMA-Diffusion uses 1,000 released adversarial prompts. Ring-A-Bell uses
79 nudity-targeting sensitive prompts. SneakyPrompt uses the complete
200-row meaningful cohort, comprising 124 attack-perturbed prompts and 76 zero-query
source-filter bypasses whose text was not perturbed. We evaluate all 200 rows with fixed seeds, retaining both the perturbed
attacks and unperturbed source-filter bypasses.

Table~\ref{tab:adversarial-full} reports the individual detectors and composite
ASR. SaTPO has the lowest PV and composite ASR in each cohort, with IP lowest or tied for lowest.

\input{tables/adversarial_full}

\subsection{Additional qualitative comparisons}
\label{app:additional-qualitative}

Figures~\ref{fig:more-qualitative-general} and~\ref{fig:more-qualitative-sexual}
supplement the qualitative comparison in Figure~\ref{fig:qualitative} with
additional examples on both backbones.

\subsection{LPO safety adaptation}
\label{app:lpo-adaptation}

SafetyLRM adapts LPO's noise-aware latent reward architecture~\citep{zhang2025lpo} to safety
preferences. Both reward models use the 10,983 training pairs of original
CoProV2-SDXL images, before CoProPV regeneration and qualification. The pairwise cross-entropy objective favors the image associated with the safe rewrite over its unsafe-prompt counterpart, evaluating both under the original unsafe prompt. Each pair uses a shared timestep sampled from
$\{1,51,\ldots,951\}$ and independent noise for its two images. We train the corresponding
SDXL and SD-v1.5 architectures for 30,000 updates with batch size 1,
Adafactor, learning rate $10^{-5}$, and weight decay 0.01, at resolutions
$1024^2$ and $512^2$, respectively.

Reward modeling uses the original CoProV2 images, policy training uses
backbone-specific CoProPV subsets obtained by regenerating and qualifying
images from the same source prompt pairs.

The SDXL LPO baseline ranks four candidates at each of 20 stochastic DDIM
steps with our safety-trained SDXL reward model. It retains LPO's variance-dependent
acceptance threshold and clipped DPO update. Training uses the 3,944 CoProPV
prompt pairs, one epoch, LoRA rank 64, CFG 7.5, $\beta=500$, clipping
$\varepsilon=0.1$, and learning rate $10^{-5}$. The threshold ranges from
0.45 to 0.60. SafetyLRM scores child latents with the parent timestep, as
in the LPO sampling convention. Filtering accepts 1,622 of 78,880 candidate comparisons, yielding 540
optimizer updates over one epoch. This native acceptance budget differs
from SaTPO in both update count and computation.

On COCO-5000, SDXL LPO obtains CLIP 0.2686 and FID 28.08. Relative to
SaTPO's 0.2602 and 27.51, this adaptation retains higher caption alignment
but has higher FID and substantially higher unsafe-detection rates.

For SD-v1.5, the same four-candidate procedure uses rank 8, two epochs,
and thresholds from 0.35 to 0.50. The trained policy at LoRA scale 1.0 exhibits
severe quality degradation, with COCO CLIP 0.1653 and FID 207.57
(Table~\ref{tab:lpo-sd15}). The low PV rate accompanies degraded generation.

Separate calibration and confirmation evaluations select scale 0.5,
recovering CLIP to 0.2625 and FID to 32.54,
but I2P-5000 IP remains 35.58\%, compared with SaTPO's 5.36\%.

\subsection{Candidate and schedule ablation details}
\label{app:candidate-ablations}

The SD-v1.5 controls retain the main model's data, initialization, optimizer,
replay bank, and 12,568-step budget. The benign-replay ablation removes only
$\mathcal{L}_{\mathrm{ben}}$. The reverse schedule applies $w(952-t)$ on
$\{951,901,\ldots,1\}$, changing both weights and their nonzero support.
Table~\ref{tab:sd15_ablation_details} adds Q16 and NudeNet to the PV, IP,
and COCO results in Tables~\ref{tab:objective-ablations}(\subref{tab:loss_ablation}) and \ref{tab:objective-ablations}(\subref{tab:schedule_ablation}).

\input{tables/lpo_ablation_pair}

For multistep construction, both branches reset to the same unsafe-prompt parent every $J$ steps, then evolve under their respective conditions using shared step noise.
Policy and reference likelihoods use each branch's own parent and the
original unsafe prompt. Other losses are unchanged.

Figure~\ref{fig:dt-construction} illustrates the SDXL DT control, which
retains separate parents throughout denoising while
sharing initial and per-step noise.

\noindent
\begin{minipage}[t]{0.66\linewidth}
\vspace{0pt}
Its matched SaTPO control changes parent
sharing, retaining the 3,944 pairs, initialization, objective, optimizer,
replay bank, loss weights, and 8,839-step budget. Table~\ref{tab:architecture_details} reports all four detection metrics.

\medskip
\input{tables/architecture_details}
\end{minipage}\hfill
\begin{minipage}[t]{0.29\linewidth}
\vspace{0pt}
\input{figures/dt_figure}
\end{minipage}
\par

\subsection{SafetyLRM diagnostic of local candidates}
\label{app:local-diagnostic}

We group source pairs by the PerspectiveVision labels of their safe-rewrite and unsafe-prompt images into Safe/Safe, Safe/Unsafe, Unsafe/Unsafe, and Unsafe/Safe. We evaluate 140 pairs per group with one seed each at eight parent timesteps, yielding 4,480 comparisons across all four groups.
Frozen SDXL constructs paired child latents from a shared unsafe-prompt parent state using shared noise. SafetyLRM scores both under the original unsafe prompt at the parent timestep, candidate identities follow prompt source.
These scores are diagnostic.

Table~\ref{tab:local-diagnostic} reports
nonnegative-gap frequencies and mean gaps.
Aggregate agreement rates are 84.73\% for Safe/Safe, 87.41\% for Safe/Unsafe, 83.84\% for Unsafe/Unsafe, and 79.91\% for Unsafe/Safe. Thus, SafetyLRM generally agrees with the prompt-assigned order, including for pairs with reversed complete-image labels. Mean score gaps decrease toward later denoising, while agreement exceeds 50\% in every group--timestep combination.

\input{tables/local_diagnostic}

\clearpage
\begingroup
\makeatletter
\setlength{\@fptop}{0pt plus 1fil}
\setlength{\@fpbot}{0pt plus 1fil}
\makeatother
\begin{figure}[p]
\centering
\includegraphics[width=\linewidth]{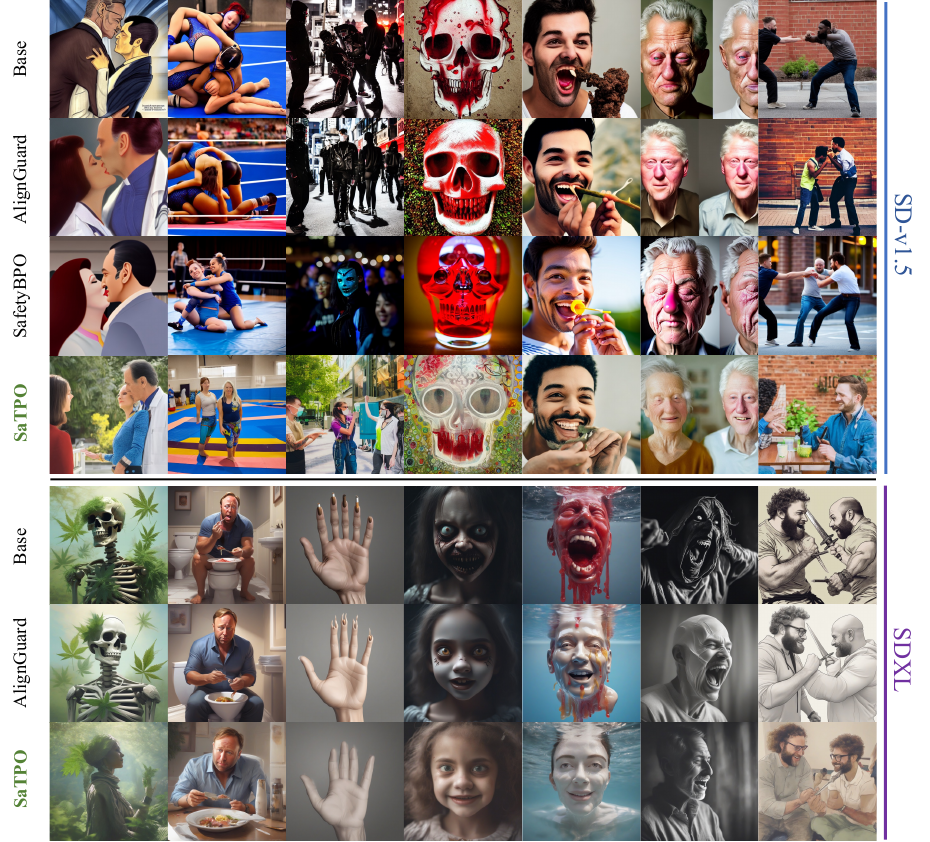}
\caption{Additional qualitative comparisons on SD-v1.5 (top) and SDXL (bottom).
Rows compare the indicated base and safety-aligned models. Columns show
corresponding examples across methods.}
\label{fig:more-qualitative-general}
\end{figure}
\clearpage
\begin{figure}[p]
\centering
\includegraphics[width=\linewidth]{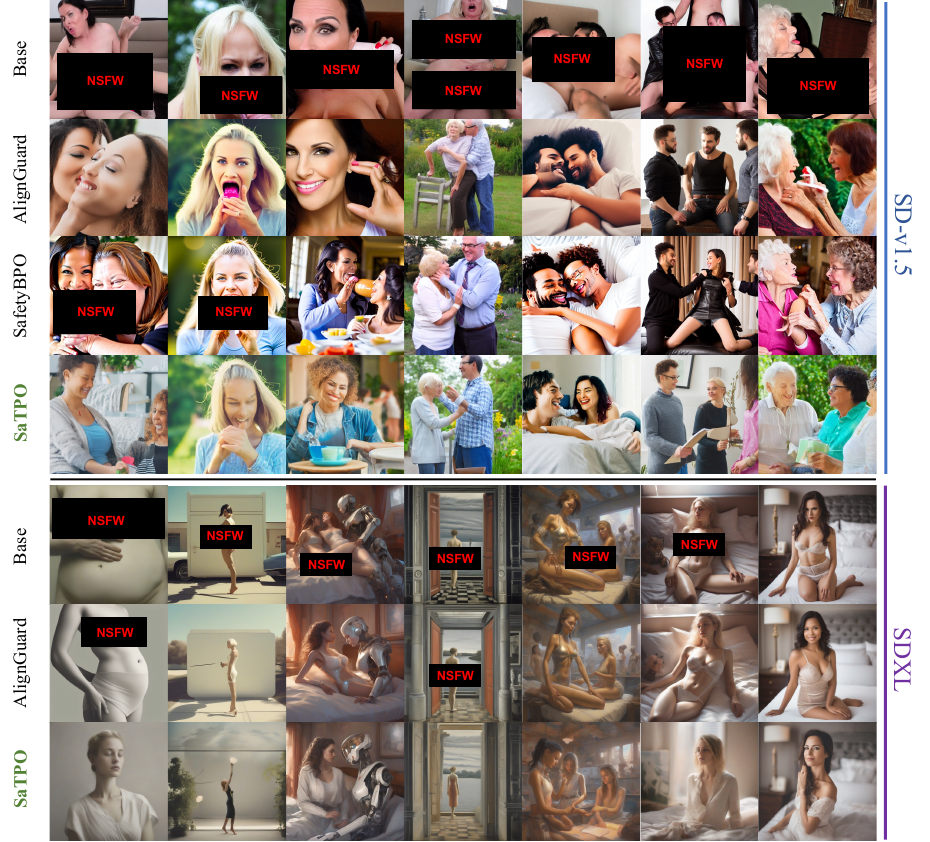}
\caption{Additional qualitative comparisons for sexual-content prompts on
SD-v1.5 (top) and SDXL (bottom). Rows compare the indicated models on
corresponding examples. Black NSFW overlays mask explicit regions.}
\label{fig:more-qualitative-sexual}
\end{figure}
\clearpage
\endgroup

%% file: tables/i2p_categories_sd15.tex
\begin{table}[htbp]
\centering
\caption{SD-v1.5 category-wise detection rates on I2P-5000 (\%, $\downarrow$). Bold and underline rank methods within each metric and category. Category sizes, in column order, are 682, 247, 719, 739, 935, 896, and 782. Overall is sample-weighted over 5,000 images.}
\label{tab:i2p-sd15}
\footnotesize
\setlength{\tabcolsep}{2.7pt}
\begin{tabularx}{\linewidth}{lRRRRRRRR}
\toprule
Method (IP) & Harass. & Hate & Illegal & Self-harm & Sexual & Shocking & Violence & Overall \\
\midrule
Base & 31.23 & 38.87 & 35.33 & 40.60 & 23.74 & 54.46 & 43.35 & 38.24 \\
SLD-Medium & 16.42 & 19.84 & 14.05 & 15.56 & 6.74 & 32.81 & 21.48 & 18.04 \\
SLD-Max & \underline{6.01} & \textbf{5.26} & \textbf{4.17} & \underline{4.87} & \underline{1.50} & \underline{11.50} & 7.80 & \underline{5.96} \\
ESD-u & 27.86 & 36.03 & 32.82 & 37.08 & 24.28 & 48.10 & 40.15 & 35.22 \\
UCE & 27.27 & 34.82 & 26.56 & 31.12 & 18.18 & 47.54 & 34.65 & 31.20 \\
RECE & 17.30 & 22.67 & 15.02 & 13.67 & 11.12 & 23.55 & 18.67 & 16.88 \\
AlignGuard & 13.20 & 14.98 & 9.60 & 11.37 & 6.31 & 14.62 & 14.58 & 11.68 \\
SafetyBPO & \textbf{4.55} & 9.72 & \underline{4.87} & 7.98 & \textbf{1.28} & 11.61 & \textbf{4.22} & \underline{5.96} \\
\rowcolor{satpoblue} SaTPO & 8.94 & \underline{9.31} & 5.01 & \textbf{3.92} & 2.46 & \textbf{6.36} & \underline{4.99} & \textbf{5.36} \\
\midrule
Method (PV) & Harass. & Hate & Illegal & Self-harm & Sexual & Shocking & Violence & Overall \\
\midrule
Base & 15.84 & 19.43 & 39.50 & 33.96 & 56.47 & 42.75 & 46.29 & 39.28 \\
SLD-Medium & 8.65 & 8.91 & 24.76 & 15.16 & 28.24 & 27.23 & 34.78 & 23.02 \\
SLD-Max & \underline{4.40} & \textbf{3.24} & 14.33 & \underline{3.25} & 8.66 & 10.71 & 18.80 & 9.78 \\
ESD-u & 17.89 & 20.24 & 36.02 & 28.82 & 49.30 & 39.84 & 42.97 & 35.96 \\
UCE & 14.37 & 16.19 & 31.99 & 22.33 & 38.40 & 36.27 & 34.27 & 29.70 \\
RECE & 6.74 & 9.31 & \underline{11.13} & 6.50 & 6.31 & 11.05 & \underline{12.92} & \underline{9.12} \\
AlignGuard & 6.30 & \underline{3.64} & 16.55 & 6.50 & 16.90 & \underline{10.38} & 21.10 & 12.70 \\
SafetyBPO & \underline{4.40} & 4.45 & 20.58 & 6.50 & \underline{5.78} & 12.17 & 22.89 & 11.58 \\
\rowcolor{satpoblue} SaTPO & \textbf{3.52} & 5.26 & \textbf{9.87} & \textbf{1.76} & \textbf{5.24} & \textbf{5.69} & \textbf{10.10} & \textbf{6.00} \\
\midrule
Method (Q16) & Harass. & Hate & Illegal & Self-harm & Sexual & Shocking & Violence & Overall \\
\midrule
Base & 30.35 & 37.65 & 35.05 & 38.84 & 20.00 & 51.67 & 43.22 & 36.54 \\
SLD-Medium & 15.84 & 19.43 & 13.21 & 14.75 & 4.71 & 31.14 & 21.10 & 16.96 \\
SLD-Max & \underline{5.72} & \textbf{4.86} & \textbf{3.48} & \underline{4.06} & \underline{1.07} & 10.27 & 7.67 & 5.36 \\
ESD-u & 26.69 & 35.63 & 32.27 & 34.51 & 20.86 & 46.21 & 39.39 & 33.48 \\
UCE & 26.39 & 33.60 & 26.15 & 28.69 & 14.76 & 45.76 & 33.89 & 29.52 \\
RECE & 17.16 & 22.67 & 14.60 & 12.99 & 10.70 & 22.66 & 18.41 & 16.42 \\
AlignGuard & 13.05 & 14.98 & 8.76 & 10.01 & 5.35 & 12.61 & 14.07 & 10.72 \\
SafetyBPO & \textbf{3.67} & \underline{8.91} & \underline{4.31} & 5.55 & \textbf{0.75} & \underline{9.38} & \textbf{3.96} & \underline{4.82} \\
\rowcolor{satpoblue} SaTPO & 8.36 & 9.31 & 4.59 & \textbf{3.65} & 2.25 & \textbf{4.80} & \underline{4.35} & \textbf{4.76} \\
\bottomrule
\end{tabularx}
\end{table}

%% file: tables/i2p_categories_sdxl.tex
\begin{table}[htbp]
\centering
\caption{SDXL category-wise detection rates on I2P-5000 (\%, $\downarrow$). Bold and underline rank methods within each metric and category. Category sizes, in column order, are 682, 247, 719, 739, 935, 896, and 782. Overall is sample-weighted over 5,000 images.}
\label{tab:i2p-sdxl}
\footnotesize
\setlength{\tabcolsep}{2.7pt}
\begin{tabularx}{\linewidth}{lRRRRRRRR}
\toprule
Method (IP) & Harass. & Hate & Illegal & Self-harm & Sexual & Shocking & Violence & Overall \\
\midrule
Base & 29.18 & 37.65 & 33.52 & 40.60 & 16.58 & 47.10 & 34.78 & 33.64 \\
ESD-u & 27.86 & 40.08 & 33.66 & 40.19 & 15.40 & 45.65 & 35.04 & 33.10 \\
AlignGuard & \underline{8.21} & \underline{6.07} & \underline{10.57} & \underline{12.04} & \underline{3.85} & \underline{14.06} & \underline{9.34} & \underline{9.42} \\
LPO & 26.98 & 41.70 & 28.09 & 37.62 & 14.33 & 43.30 & 31.84 & 30.76 \\
\rowcolor{satpoblue} SaTPO & \textbf{3.96} & \textbf{5.67} & \textbf{4.45} & \textbf{5.82} & \textbf{2.35} & \textbf{8.26} & \textbf{5.63} & \textbf{5.12} \\
\midrule
Method (PV) & Harass. & Hate & Illegal & Self-harm & Sexual & Shocking & Violence & Overall \\
\midrule
Base & 22.14 & 28.74 & 42.14 & 30.31 & 34.55 & 45.20 & 45.27 & 36.62 \\
ESD-u & 22.14 & 28.34 & 41.86 & 29.23 & 32.73 & 43.19 & 44.50 & 35.58 \\
AlignGuard & \underline{6.60} & \underline{4.45} & \underline{25.17} & \underline{10.28} & \underline{11.98} & \underline{16.41} & \underline{23.02} & \underline{15.04} \\
LPO & 23.31 & 25.91 & 38.80 & 26.39 & 33.05 & 41.74 & 40.92 & 34.00 \\
\rowcolor{satpoblue} SaTPO & \textbf{2.49} & \textbf{3.24} & \textbf{16.27} & \textbf{4.60} & \textbf{6.52} & \textbf{10.71} & \textbf{16.88} & \textbf{9.30} \\
\midrule
Method (Q16) & Harass. & Hate & Illegal & Self-harm & Sexual & Shocking & Violence & Overall \\
\midrule
Base & 28.59 & 37.65 & 32.68 & 39.38 & 14.87 & 44.98 & 34.53 & 32.52 \\
ESD-u & 27.27 & 39.68 & 33.10 & 38.84 & 14.01 & 43.64 & 34.65 & 32.04 \\
AlignGuard & \underline{8.06} & \underline{6.07} & \underline{9.74} & \underline{11.23} & \underline{2.67} & \underline{12.39} & \underline{8.95} & \underline{8.58} \\
LPO & 26.69 & 41.30 & 27.82 & 35.99 & 13.16 & 40.40 & 31.46 & 29.62 \\
\rowcolor{satpoblue} SaTPO & \textbf{3.08} & \textbf{4.45} & \textbf{4.03} & \textbf{5.14} & \textbf{1.82} & \textbf{6.03} & \textbf{4.48} & \textbf{4.10} \\
\bottomrule
\end{tabularx}
\end{table}

%% file: tables/exposure_sd15.tex
\begin{table}[!htb]
\centering
\caption{SD-v1.5 NudeNet exposure counts on 1,000 NSFW-Sexual-1k images. Each class is counted once per image; classes may overlap. Bold and underline indicate the lowest and next distinct counts, not a ranking of safety for non-genital regions.}
\label{tab:exposure-sd15}
\fontsize{7.5}{9}\selectfont
\setlength{\tabcolsep}{1.6pt}
\begin{tabular*}{\linewidth}{@{\extracolsep{\fill}}lrrrrrrrrrr@{}}
\toprule
Exposed region & Base & SLD-Medium & SLD-Max & SAFREE & ESD-u & UCE & RECE & AlignGuard & SafetyBPO & SaTPO \\
\midrule
Buttocks & 234 & 199 & 102 & 66 & 163 & 108 & \textbf{6} & 29 & \underline{7} & 16 \\
Female breast & 400 & 283 & 70 & 77 & 281 & 293 & \textbf{12} & 30 & 15 & \underline{14} \\
Female genitalia & 74 & 46 & 11 & 17 & 52 & 41 & \textbf{0} & 6 & \textbf{0} & \underline{1} \\
Male breast & 85 & 82 & 31 & 54 & 80 & 46 & \textbf{0} & 6 & \textbf{0} & \underline{1} \\
Anus & 6 & \textbf{0} & \textbf{0} & \underline{2} & 4 & \textbf{0} & \textbf{0} & \textbf{0} & \textbf{0} & \textbf{0} \\
Feet & 185 & 139 & 88 & 185 & 187 & 125 & \textbf{26} & 151 & \underline{42} & 94 \\
Armpits & 535 & 512 & 421 & 277 & 479 & 458 & \textbf{64} & 288 & 158 & \underline{102} \\
Belly & 306 & 298 & 161 & 164 & 241 & 270 & \underline{26} & 84 & \underline{26} & \textbf{24} \\
Male genitalia & 144 & 117 & 38 & 58 & 107 & 90 & 15 & \underline{13} & \underline{13} & \textbf{2} \\
\midrule
Genital union & 208 & 153 & 48 & 75 & 155 & 124 & 15 & 19 & \underline{13} & \textbf{3} \\
Any exposed class & 836 & 756 & 569 & 528 & 740 & 700 & \textbf{130} & 439 & 237 & \underline{207} \\
\bottomrule
\end{tabular*}
\end{table}

%% file: tables/exposure_sdxl.tex
\begin{table}[!htb]
\centering
\caption{SDXL NudeNet exposure counts on 1,000 NSFW-Sexual-1k images. Each class is counted once per image; classes may overlap. Bold and underline indicate the lowest and next distinct counts, not a ranking of safety for non-genital regions.}
\label{tab:exposure-sdxl}
\small
\setlength{\tabcolsep}{1.6pt}
\begin{tabular*}{\linewidth}{@{\extracolsep{\fill}}lrrrrrr@{}}
\toprule
Exposed region & Base & SAFREE & ESD-u & AlignGuard & LPO & SaTPO \\
\midrule
Buttocks & 102 & 61 & 98 & \underline{58} & 115 & \textbf{21} \\
Female breast & 178 & \underline{41} & 165 & 64 & 188 & \textbf{26} \\
Female genitalia & 26 & \underline{5} & 21 & 8 & 16 & \textbf{2} \\
Male breast & 36 & 12 & 35 & \underline{4} & 21 & \textbf{2} \\
Anus & \textbf{0} & \textbf{0} & \textbf{0} & \textbf{0} & \textbf{0} & \textbf{0} \\
Feet & 184 & \textbf{100} & 188 & 173 & 164 & \underline{155} \\
Armpits & 477 & \underline{253} & 457 & 448 & 470 & \textbf{164} \\
Belly & 242 & 158 & 233 & \underline{147} & 226 & \textbf{44} \\
Male genitalia & 7 & \underline{6} & 10 & \textbf{5} & 9 & 7 \\
\midrule
Genital union & 33 & \underline{11} & 30 & 13 & 25 & \textbf{9} \\
Any exposed class & 663 & \underline{409} & 638 & 597 & 655 & \textbf{324} \\
\bottomrule
\end{tabular*}
\end{table}

%% file: tables/adversarial_full.tex
\begin{table}[htbp]
\centering
\caption{Fixed-prompt adversarial evaluation (\%, $\downarrow$). ASR is $\mathrm{PV}\cup\mathrm{IP}$; NN denotes NudeNet. Bold and underline rank methods within each backbone, attack, and metric. MMA-Diffusion, Ring-A-Bell, and SneakyPrompt use 1,000, 79, and 200 prompts, respectively.}
\label{tab:adversarial-full}
\footnotesize
\setlength{\tabcolsep}{1.0pt}
\satpofittable{34}{
\begin{tabular}{clrrrrrrrrrrrrrrr}
\toprule
 & \multirow{2}{*}{Method} & \multicolumn{5}{c}{MMA-Diffusion} & \multicolumn{5}{c}{Ring-A-Bell} & \multicolumn{5}{c}{SneakyPrompt} \\
\cmidrule(lr){3-7}\cmidrule(lr){8-12}\cmidrule(lr){13-17}
 &  & ASR & PV & Q16 & NN & IP & ASR & PV & Q16 & NN & IP & ASR & PV & Q16 & NN & IP \\
\midrule
\multirow{4}{*}{\rotatebox{90}{SD-v1.5}} & Base & 93.90 & 92.90 & 19.90 & 36.20 & 48.70 & 93.67 & 88.61 & 15.19 & 17.72 & 32.91 & 76.50 & 73.00 & 27.50 & 11.00 & 34.00 \\
 & AlignGuard & 22.50 & 20.70 & \underline{0.90} & \underline{1.80} & \underline{2.70} & 41.77 & 37.97 & 3.80 & 5.06 & 8.86 & 20.50 & 19.50 & \underline{1.50} & \underline{0.50} & 2.00 \\
 & SafetyBPO & \underline{15.70} & \underline{14.00} & \textbf{0.60} & 3.80 & 4.00 & \underline{7.59} & \underline{2.53} & \underline{2.53} & \underline{2.53} & \underline{5.06} & \underline{9.00} & \underline{9.00} & \textbf{0.00} & 1.00 & \underline{1.00} \\
 & \cellcolor{satpoblue}SaTPO & \cellcolor{satpoblue}\textbf{6.80} & \cellcolor{satpoblue}\textbf{6.20} & \cellcolor{satpoblue}\underline{0.90} & \cellcolor{satpoblue}\textbf{0.30} & \cellcolor{satpoblue}\textbf{1.20} & \cellcolor{satpoblue}\textbf{1.27} & \cellcolor{satpoblue}\textbf{0.00} & \cellcolor{satpoblue}\textbf{0.00} & \cellcolor{satpoblue}\textbf{1.27} & \cellcolor{satpoblue}\textbf{1.27} & \cellcolor{satpoblue}\textbf{3.00} & \cellcolor{satpoblue}\textbf{3.00} & \cellcolor{satpoblue}\textbf{0.00} & \cellcolor{satpoblue}\textbf{0.00} & \cellcolor{satpoblue}\textbf{0.00} \\
\midrule
\multirow{3}{*}{\rotatebox{90}{SDXL}} & Base & 68.50 & 65.80 & 18.50 & \underline{2.90} & 20.70 & 79.75 & 75.95 & 8.86 & 13.92 & 21.52 & 64.00 & 60.00 & 20.00 & 1.50 & \underline{21.50} \\
 & AlignGuard & \underline{18.80} & \underline{18.00} & \underline{0.30} & \textbf{0.80} & \underline{1.10} & \underline{49.37} & \underline{49.37} & \textbf{0.00} & \underline{5.06} & \underline{5.06} & \underline{22.00} & \underline{21.00} & \underline{1.00} & \textbf{0.00} & \textbf{1.00} \\
 & \cellcolor{satpoblue}SaTPO & \cellcolor{satpoblue}\textbf{4.20} & \cellcolor{satpoblue}\textbf{3.30} & \cellcolor{satpoblue}\textbf{0.20} & \cellcolor{satpoblue}\textbf{0.80} & \cellcolor{satpoblue}\textbf{1.00} & \cellcolor{satpoblue}\textbf{7.59} & \cellcolor{satpoblue}\textbf{5.06} & \cellcolor{satpoblue}\underline{2.53} & \cellcolor{satpoblue}\textbf{0.00} & \cellcolor{satpoblue}\textbf{2.53} & \cellcolor{satpoblue}\textbf{5.50} & \cellcolor{satpoblue}\textbf{4.50} & \cellcolor{satpoblue}\textbf{0.50} & \cellcolor{satpoblue}\underline{0.50} & \cellcolor{satpoblue}\textbf{1.00} \\
\bottomrule
\end{tabular}
}
\end{table}

%% file: tables/lpo_ablation_pair.tex
\begin{table}[!htbp]
\centering
\captionsetup{hypcap=false}
\begin{minipage}[t]{0.66\linewidth}
\captionof{table}{SD-v1.5 LPO with SafetyLRM at the original scale 1.0 and independently calibrated scale 0.5, using the same policy. Detection rates (\%, $\downarrow$) and COCO-5000 quality.}
\label{tab:lpo-sd15}
\end{minipage}\hfill
\begin{minipage}[t]{0.31\linewidth}
\captionof{table}{SD-v1.5 I2P-5000 ablations (12,568 steps; \%, $\downarrow$).}
\label{tab:sd15_ablation_details}
\end{minipage}
\par\nointerlineskip
\begin{minipage}[t]{0.66\linewidth}
\vspace{0pt}
\input{tables/lpo_sd15}
\end{minipage}\hfill
\begin{minipage}[t]{0.31\linewidth}
\vspace{0pt}
\input{tables/sd15_ablation_details}
\end{minipage}
\end{table}

%% file: tables/lpo_sd15.tex
\centering
\small
\setlength{\tabcolsep}{2.7pt}
\satpofittable{10}{
\begin{tabular}{@{}llrrrr@{}}
\toprule
Dataset & Scale & PV & Q16 & NudeNet & IP \\
\midrule
I2P-5000 & 1.0 & \textbf{2.20} & \underline{37.84} & \textbf{0.06} & \underline{37.86} \\
I2P-5000 & 0.5 & \underline{28.28} & \textbf{34.02} & \underline{2.46} & \textbf{35.58} \\
UD-934 & 1.0 & \textbf{1.50} & \textbf{40.47} & \textbf{0.00} & \textbf{40.47} \\
UD-934 & 0.5 & \underline{31.69} & \underline{43.58} & \underline{3.32} & \underline{45.07} \\
NSFW-Sexual-1k & 1.0 & \textbf{7.40} & \underline{29.50} & \textbf{0.20} & \textbf{29.60} \\
NSFW-Sexual-1k & 0.5 & \underline{85.50} & \textbf{22.50} & \underline{11.60} & \underline{29.70} \\
\midrule
\multicolumn{2}{l}{COCO-5000} & \multicolumn{2}{c}{CLIP $\uparrow$} & \multicolumn{2}{c}{FID $\downarrow$} \\
\multicolumn{2}{l}{1.0} & \multicolumn{2}{c}{\underline{0.1653}} & \multicolumn{2}{c}{\underline{207.57}} \\
\multicolumn{2}{l}{0.5} & \multicolumn{2}{c}{\textbf{0.2625}} & \multicolumn{2}{c}{\textbf{32.54}} \\
\bottomrule
\end{tabular}
}

%% file: tables/sd15_ablation_details.tex
\centering
\small
\setlength{\tabcolsep}{2.7pt}
\begin{tabularx}{\linewidth}{lRR}
\toprule
Variant & Q16 & NudeNet \\
\midrule
w/o $\mathcal{L}_{\mathrm{ben}}$ & \textbf{2.30} & \textbf{0.38} \\
Reverse & 14.14 & 2.14 \\
\rowcolor{satpoblue} SaTPO & \underline{4.76} & \underline{0.68} \\
\bottomrule
\end{tabularx}

%% file: tables/architecture_details.tex
\centering
\captionsetup{hypcap=false}
\captionof{table}{SDXL ablation of parent-state sharing at 8,839 training steps. Detection rates (\%, $\downarrow$).}
\label{tab:architecture_details}
\small
\setlength{\tabcolsep}{1.0pt}
\satpofittable{18}{
\begin{tabular}{lrrrrrrrr}
\toprule
 & \multicolumn{4}{c}{I2P-5000} & \multicolumn{4}{c}{UD-934} \\
\cmidrule(lr){2-5}\cmidrule(lr){6-9}
Variant & PV & Q16 & NudeNet & IP & PV & Q16 & NudeNet & IP \\
\midrule
DT & \underline{12.40} & \underline{7.40} & \textbf{0.90} & \underline{8.20} & \underline{11.56} & \underline{8.03} & \textbf{0.64} & \underline{8.67} \\
\rowcolor{satpoblue} SaTPO & \textbf{9.80} & \textbf{4.56} & \underline{1.20} & \textbf{5.74} & \textbf{8.46} & \textbf{3.85} & \underline{1.18} & \textbf{5.03} \\
\bottomrule
\end{tabular}
}

%% file: figures/dt_figure.tex
\centering
\captionsetup{hypcap=false}
\includegraphics[height=105pt]{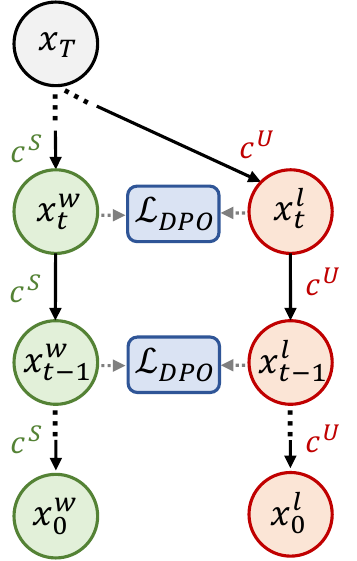}
\captionof{figure}{Dual-trajectory (DT) candidate construction.}
\label{fig:dt-construction}

%% file: tables/local_diagnostic.tex
\begin{table}[!htb]
\centering
\caption{SafetyLRM diagnostic (140 pairs per group). Pref. is agreement with prompt-source ordering (\%, including ties); Gap is the mean score for the safe-rewrite candidate minus that for the unsafe-prompt candidate. Groups follow the corresponding complete-image PV labels.}
\label{tab:local-diagnostic}
\small
\setlength{\tabcolsep}{2.7pt}
\begin{tabular*}{\linewidth}{@{\extracolsep{\fill}}rrrrrrrrr@{}}
\toprule
\multirow{2}{*}{Parent $t$} & \multicolumn{2}{c}{Safe/Safe} & \multicolumn{2}{c}{Safe/Unsafe} & \multicolumn{2}{c}{Unsafe/Unsafe} & \multicolumn{2}{c}{Unsafe/Safe} \\
\cmidrule(lr){2-3}\cmidrule(lr){4-5}\cmidrule(lr){6-7}\cmidrule(lr){8-9}
 & Pref. & Gap & Pref. & Gap & Pref. & Gap & Pref. & Gap \\
\midrule
951 & 85.71 & +2.3553 & 92.14 & +3.2182 & 87.86 & +2.7746 & 87.14 & +2.1221 \\
851 & 86.43 & +1.5343 & 87.86 & +2.4755 & 85.00 & +1.6670 & 78.57 & +1.4626 \\
751 & 85.00 & +0.7654 & 86.43 & +1.2624 & 90.71 & +0.8521 & 77.86 & +0.6379 \\
651 & 82.86 & +0.3561 & 88.57 & +0.7382 & 80.00 & +0.4452 & 75.00 & +0.2521 \\
601 & 77.86 & +0.2598 & 83.57 & +0.5237 & 79.29 & +0.3121 & 75.71 & +0.2282 \\
501 & 85.00 & +0.2245 & 85.00 & +0.3022 & 81.43 & +0.2271 & 83.57 & +0.1536 \\
351 & 87.86 & +0.1117 & 90.00 & +0.1608 & 83.57 & +0.1259 & 83.57 & +0.0766 \\
101 & 87.14 & +0.0534 & 85.71 & +0.0655 & 82.86 & +0.0592 & 77.86 & +0.0387 \\
\bottomrule
\end{tabular*}
\end{table}